\documentclass[10pt,twocolumn,letterpaper]{article}
\usepackage{titling}
\usepackage{iccv}
\usepackage{times}
\usepackage{epsfig}
\usepackage{graphicx}
\usepackage{amsmath}
\usepackage{amssymb}
\usepackage{marvosym}
\usepackage{multirow}
\usepackage{bm}
\usepackage{ifthen}
\usepackage{threeparttable}

\usepackage[pagebackref=true,breaklinks=true,letterpaper=true,colorlinks,bookmarks=false]{hyperref}

\iccvfinalcopy 

\def\iccvPaperID{} 

\begin{document}

{
{\title{Learning Non-Local Spatial-Angular Correlation for Light Field \\Image Super-Resolution}

\author{Zhengyu Liang$^{1}$, Yingqian Wang$^{1}$, Longguang Wang$^{2}$, Jungang Yang$^{1}$\textsuperscript{\Letter}, Shilin Zhou$^{1}$, Yulan Guo$^{1}$\\
$^{1}$National University of Defense Technology, $^{2}$Aviation University of Air Force\\
{\tt\small \{zyliang, yangjungang\}@nudt.edu.cn}}
}

\maketitle
}

\begin{abstract}
Exploiting spatial-angular correlation is crucial to light field (LF) image super-resolution (SR), but is highly challenging due to its non-local property caused by the disparities among LF images. Although many deep neural networks (DNNs) have been developed for LF image SR and achieved continuously improved performance, existing methods cannot well leverage the long-range spatial-angular correlation and thus suffer a significant performance drop when handling scenes with large disparity variations. In this paper, we propose a simple yet effective method to learn the non-local spatial-angular correlation for LF image SR. In our method, we adopt the epipolar plane image (EPI) representation to project the 4D spatial-angular correlation onto multiple 2D EPI planes, and then develop a Transformer network with repetitive self-attention operations to learn the spatial-angular correlation by modeling the dependencies between each pair of EPI pixels. Our method can fully incorporate the information from all angular views while achieving a global receptive field along the epipolar line. We conduct extensive experiments with insightful visualizations to validate the effectiveness of our method. Comparative results on five public datasets show that our method not only achieves state-of-the-art SR performance but also performs robust to disparity variations. 
Code is publicly available at {\url{https://github.com/ZhengyuLiang24/EPIT}}.
\end{abstract}

\section{Introduction}
\label{Sec_Introduction}
Light field (LF) cameras record both intensity and directions of light rays, and enable various applications such as depth perception \cite{jin2022occlusion, khan2021differentiable, leistner2022towards}, view rendering \cite{attal2022learning, suhail2022light, wizadwongsa2021nex}, virtual reality \cite{choi2021neural, yu2017light}, and 3D reconstruction \cite{cai2018ray, zhang2021learning}. 
However, due to the inherent spatial-angular trade-off \cite{zhu2019revisiting}, an LF camera can either provide dense angular samplings with low-resolution (LR) sub-aperture images (SAIs), or capture high-resolution (HR) SAIs with sparse angular sampling. 
To handle this problem, many methods have been proposed to enhance the angular resolution via novel view synthesis \cite{kalantari2016learning, meng2020high, ShearedEPI}, or enhance the spatial resolution by performing LF image super-resolution (SR) \cite{LFZSSR, guo2021deep}. 
In this paper, we focus on the latter task, i.e., generating HR LF images from their LR counterparts.

\begin{figure}
    \centering
    \includegraphics[width=8.4cm]{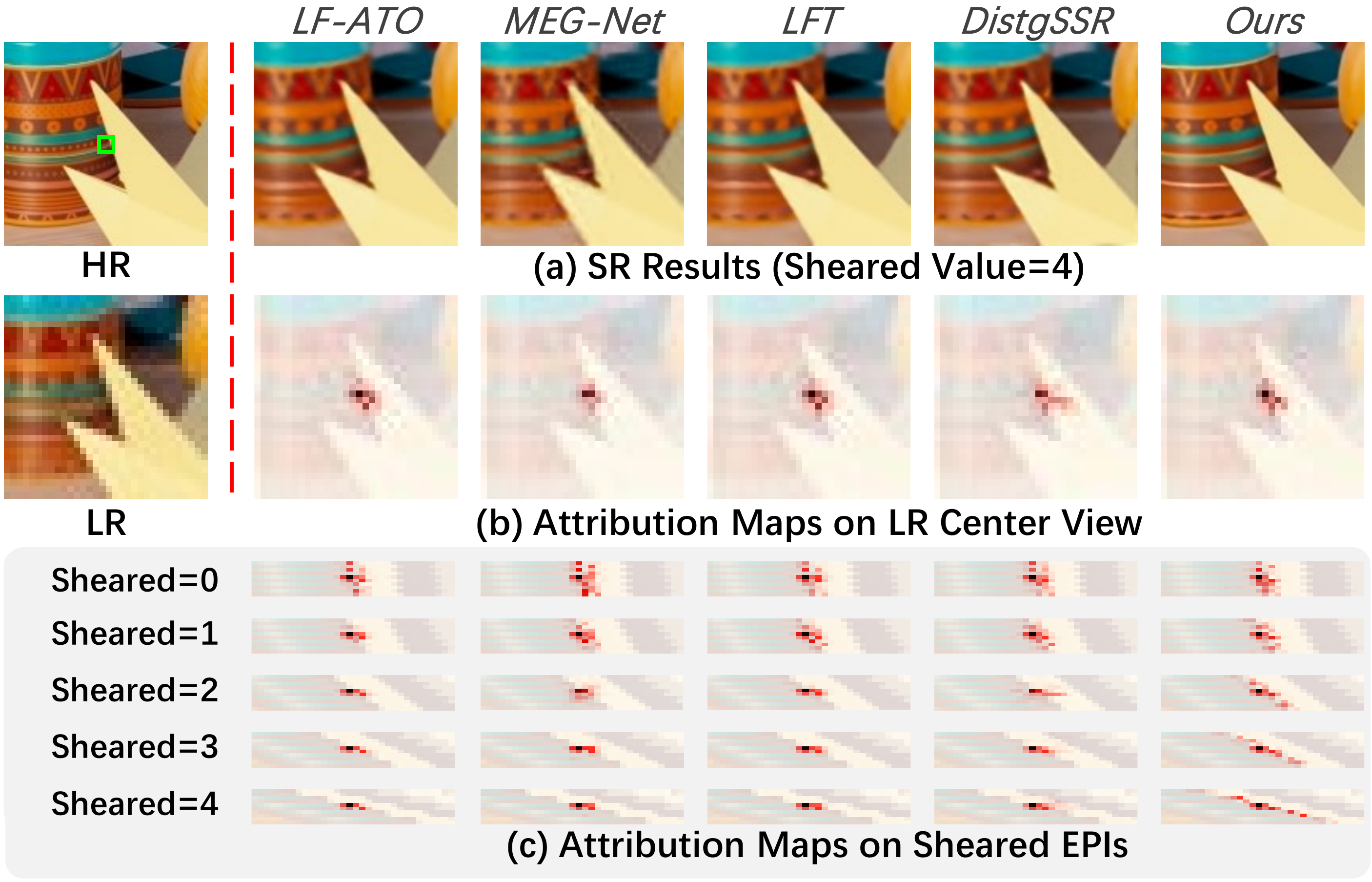}
    \caption{Visualization of 4$\times$ SR results and the corresponding attribution maps \cite{LAM} of our method and four state-of-the-art methods \cite{LFATO,LFT,LF-Distg,MegNet} under different manually sheared disparity values. Here, the patch marked by the green box in the HR image is selected as the target region, and the regions that contribute to the final SR results are highlighted in red. Our method can well exploit the non-local spatial-angular correlation and achieve superior SR performance. More examples are provided in Fig.~\ref{fig:LFSheared}.} 
    \vspace{-0.3cm}
    \label{fig:LAM}
\end{figure}

Recently, convolutional neural networks (CNNs) have been widely applied to LF image SR and demonstrated superior performance over traditional paradigms \cite{alain2018light, liang2015light, mitra2012light, rossi2018geometry, wanner2013variational}. 
To incorporate the complementary information (i.e., angular information) from different views, existing CNNs adopted various mechanisms such as adjacent-view combination \cite{LFCNN}, view-stack integration \cite{LFATO, MegNet, resLF}, bidirectional recurrent fusion \cite{LFNet}, spatial-angular disentanglement \cite{LFT, LF-Distg, LF-InterNet, LFSSR, LFSAV}, and 4D convolutions \cite{HDDRNet, meng2020high}. 
However, as illustrated in both Fig.~\ref{fig:LAM} and Sec.~\ref{sec:ShearedDisparity},  existing methods achieve promising results on LFs with small baselines, but suffer a notable performance drop when handling scenes with large disparity variations.

We attribute this performance drop to the contradictions between the local receptive field of CNNs and the requirement of incorporating \textit{non-local spatial-angular correlation} in LF image SR. That is, LF images provide multiple observations of a scene from a number of regularly distributed viewpoints, and a scene point is projected onto different but correlated spatial locations on different angular views, which is termed as the \textit{spatial-angular correlation}. Note that, the spatial-angular correlation has the non-local property since the difference between the spatial locations of two views (i.e., disparity value) depends on several factors (e.g., angular coordinates of the selected views, the depth value of the scene point, the baseline length of the LF camera, and the resolution of LF images), and can be very large in some situations. Consequently, it is appealing for LF image SR methods to incorporate complementary information from different views by exploiting the spatial-angular correlation under large disparity variations.

In this paper, we propose a simple yet effective method to learn the non-local spatial-angular correlation for LF image SR. In our method, we re-organize 4D LFs as multiple 2D epipolar plane images (EPIs) to manifest the spatial-angular correlation to the line patterns with different slopes. Then, we develop a Transformer-based network called EPIT to learn the spatial-angular correlation by modeling the dependencies between each pair of pixels on EPIs. 
 Specifically, we design a basic Transformer block to alternately process horizontal and vertical EPIs, and thus progressively incorporate the complementary information from all angular views.
 Compared to existing LF image SR methods, our method can achieve a global receptive field along the epipolar line, and thus performs robust to disparity variations.

In summary, the contributions of this work are as follows: (1) We address the importance of exploiting non-local spatial-angular correlation in LF image SR, and propose a simple yet effective method to handle this problem.
 (2) We develop a Transformer-based network to learn the non-local spatial-angular correlation from horizontal and vertical EPIs, and validate the effectiveness of our method through extensive experiments and visualizations. 
 (3) Compared to existing state-of-the-art LF image SR methods, our method achieves superior performance on public LF datasets, and is much more robust to disparity variations.

\section{Related Work}
\label{sec:related}

\subsection{LF Image SR}

LFCNN \cite{LFCNN} is the first method to adopt CNNs to learn the correspondence among stacked SAIs.
Then, it is a common practice for LF image SR networks to aggregate the complementary information from adjacent views to model the correlation in LFs. 
Yeung et al. \cite{LFSSR} designed a spatial-angular separable (SAS) convolution to approximate the 4D convolution to characterize the sub-pixel relationship of LF 4D structures. 
Wang et al. \cite{LFNet} proposed a bidirectional recurrent network to model the spatial correlation among views on horizontal and vertical baselines. 
Meng et al. \cite{HDDRNet} proposed a densely-connected network with 4D convolutions to explicitly learn the spatial-angular correlation encoded in 4D LF data. 
To further learn inherent corresponding relations in SAIs, Zhang et al. \cite{MegNet, resLF} grouped LFs into four different branches according to the specific angular directions, and used four sub-networks to model the multi-directional spatial-angular correlation.

The aforementioned networks use part of input views to super-resolve each view, and the inherent spatial-angular correlation in LF images cannot be well incorporated. 
To address this issue, Jin et al. \cite{LFATO} proposed an All-to-One framework for LF image SR, and each SAI can be individually super-resolved by combining the information from all views. 
Wang et al. \cite{LF-InterNet, LF-Distg} organized LF images into macro-pixels, and designed a disentangling mechanism to fully incorporate the angular information.
Liu et al. \cite{LF-IINet} introduced 3D convolutions based multi-view context block to exploit the correlations among all views. 
In addition, Wang et al. \cite{LF-DFnet} adopted deformable convolutions to achieve long-range information exploitation from all SAIs. 
Existing methods generally learn the local correspondence across SAIs, and ignore the importance of non-local spatial-angular correlation in LF images. 
However, due to the limited receptive field of CNNs, existing methods generally learn the local correspondence across SAIs, and ignore the importance of non-local spatial-angular correlation in LF images.

\textcolor{black}{Recently, 
Liang et al. \cite{LFT} applied Transformers to LF image SR and developed an angular Transformer and a spatial Transformer to incorporate angular information and model long-range spatial dependencies, respectively. However, since 4D LFs were organized into 2D angular patches to form the input of angular Transformers, the non-local property of spatial-angular correlations reduces the robustness of LFT to large disparity variations.
}

\subsection{Non-Local Correlation Modeling}

Non-local means \cite{NonLocalMeans} is a classical algorithm that computes the weighted mean of pixels in an image according to the self-similarity measure, and a number of studies on such non-local priors have been proposed for image restoration \cite{BM3D, singh2014super, gu2014weighted, berman2016non}, image and video SR \cite{glasner2009super, zeyde2010single, freedman2011image, yang2013fast, huang2015single}.
Then, the attention mechanism is developed as a tool to bias the most informative components of an input signal, and achieves significant performance in various computer vision tasks \cite{SEnet, chen2016attention, wang2018non, fu2019dual}.
Huang et al. \cite{CCnet} proposed novel criss-cross attention to capture contextual information from full-image dependencies in an efficient way.
Wang et al. \cite{PASSRnet, PAM} proposed a parallax attention mechanism to handle the varying disparities problem of stereo images. 
\textcolor{black}{
Wu et al. \cite{SAAN} applied attention mechanisms to 3D LF reconstruction and developed a spatial-angular attention module to learn the first-order correlation on EPIs. 
}

Recently, the attention mechanism is further generalized as Transformers \cite{vaswani2017attention} with multi-head structures and feed-forward networks. 
Transformers have inspired lots of works \cite{SwinTransformer, SwinIR, IPT, ViT} to further investigate the power of attention mechanisms in visions. 
Liu et al. \cite{liu2022video} presented a pure-Transformer method to incorporate the inherent spatial-temporal locality of videos for action recognition.
Naseer et al. \cite{naseer2021intriguing} investigated the robustness and generalizability of Transformers, and demonstrated favorable merits of Transformers over CNNs for occlusion handling. 
Shi et al. \cite{shi2022rethinking} observed that Transformers can implicitly make accurate connections for misaligned pixels, and presented a new understanding of Transformers to process spatially unaligned images.

\section{Method}
\subsection{Preliminary}
Based on the two-plane LF parameterization model \cite{LF_two_plane}, an LF image is commonly formulated as a 4D function $\mathcal{L}(u,v,h,w) \in \mathbb{R}^{U \times V \times H \times W}$, where $U$ and $V$ represent angular dimensions, $H$ and $W$ represent spatial dimensions.
The EPI sample of 4D LF is acquired with a fixed angular coordinate and a fixed spatial coordinate. 
Specifically, the horizontal EPI is obtained with constant $u$ and $h$, and the vertical EPI is obtained with constant $v$ and $w$.

\begin{figure}[t]
    \centering
    \includegraphics[height=4.5cm]{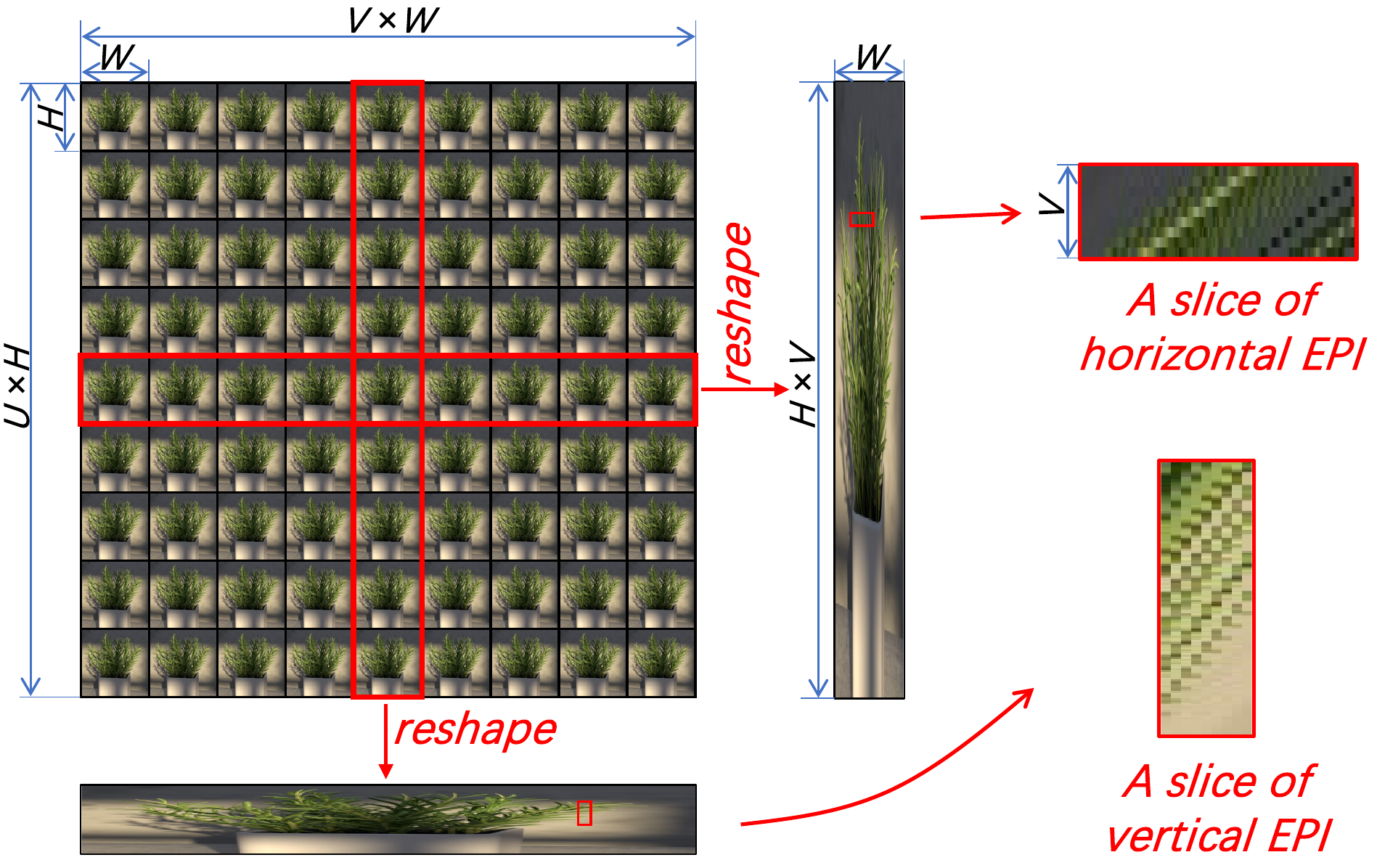}
    \caption{The SAI and EPI representations of LF images. The array of 9$\times$9 views of
    a scene \texttt{rosemary} from HCInew \cite{HCInew} dataset is used as an example for illustration. }
    \label{fig:LF_representation}
    \vspace{-0.3cm}
\end{figure}

As shown in Fig.~\ref{fig:LF_representation}, the EPIs not only record spatial structures at edges or textures, but also reflect the disparity information via line patterns of different slopes. 
Specifically, due to large disparities, the EPIs contain abundant spatial-angular correlation of LFs in a long-range way.
Therefore, we propose to explore the non-local spatial-angular correlation from horizontal and vertical EPIs for LF image SR.

\subsection{Network Design}
As shown in Fig.~\ref{fig:Overview}(a), our network takes an LR LF $\mathcal{L}_{LR} \in \mathbb{R}^{U \times V \times H \times W}$ as its input, and produces an HR LF $\mathcal{L}_{SR} \in \mathbb{R}^{U \times V \times \alpha H \times \alpha W}$, where $\alpha$ presents the upscaling factor. 
Our network consists of three stages including initial feature extraction, deep spatial-angular correlation learning, and feature upsampling.

\subsubsection{Initial Feature Extraction}
\begin{figure*}[t]
    \centering
    \includegraphics[width=17.2cm]{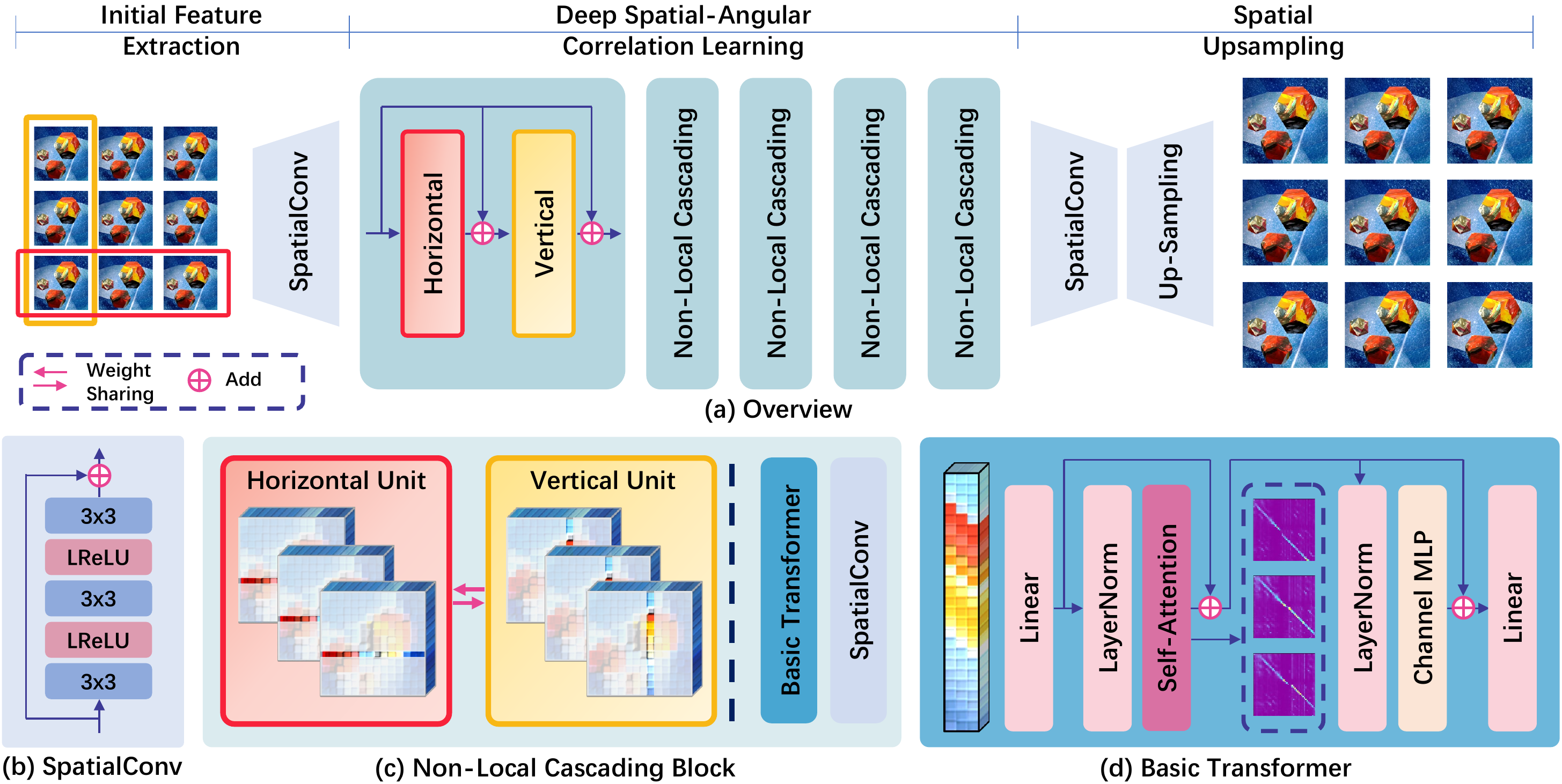}
    \caption{An overview of our proposed EPIT. Here, a 3$\times$3 LF is used as an example for illustration.}
    \label{fig:Overview}
    \vspace{-0.2cm}
\end{figure*}

As shown in Fig.~\ref{fig:Overview}(b), we follow most existing works \cite{IPT, SwinIR, MANA} to use three 3$\times$3 convolutions with LeakyReLU \cite{LeakyReLU} as a {\it SpatialConv} layer to map each SAI to a high-dimensional feature.
The initially extracted feature can be represented as ${\boldsymbol{F}}\in \mathbb{R}^{U \times V \times H \times W \times C}$, where $C$ denotes the channel dimension.

\subsubsection{Deep Spatial-Angular Correlation Learning}

\noindent{\bf Non-Local Cascading Block.}
The basic module for spatial-angular correlation learning is the {\it Non-Local Cascading} block. 
As shown in Fig.~\ref{fig:Overview}(a), each block consists of two cascaded {\it Basic-Transformer} units to separately incorporate the complementary information along the horizontal and vertical epipolar lines. 
In our method, we employed five {\it Non-Local Cascading} blocks to achieve a global perception of all angular views, and followed SwinIR \cite{SwinIR} to adopt spatial convolutions to enhance the local feature representation. 
The effectiveness of this inter-block spatial convolution is validated in Sec.~\ref{sec:ModelAnalysis}.
Note that, the weights of the two {\it Basic-Transformer} units in each block are shared to jointly learn the intrinsic properties of LFs, which is demonstrated effective in Sec.~\ref{sec:ModelAnalysis}.

As shown in Fig.~\ref{fig:Overview}(c), the initial features ${\boldsymbol F}$ can be first separately reshaped to the horizontal EPI pattern ${\boldsymbol F}_{hor} \in \mathbb{R}^{UH \times V \times W \times C}$ and the vertical EPI pattern ${\boldsymbol F}_{ver} \in \mathbb{R}^{VW \times U \times H \times C}$. 
Next, ${\boldsymbol F}_{hor}$ (or ${\boldsymbol F}_{ver}$) is fed to a {\it Basic-Transformer} unit to integrate the long-range information along the horizontal (or vertical) epipolar line.
Then, the enhanced feature $\hat{{\boldsymbol F}}_{hor}$ (or $\hat{{\boldsymbol F}}_{ver}$) is reshaped into the size of ${UV \times H \times W \times C}$, and passes through a {\it SpatialConv} layer to incorporate the spatial context information within each SAI. 
Without loss of generality, we take the vertical {\it Basic-Transformer} as an example to introduce the detail of our {\it Basic-Transformer} unit in the following texts.

\noindent{\bf Basic-Transformer Unit.}
The objective of this unit is to capture long-range dependencies along the epipolar line via Transformers. 
To leverage the powerful sequence modeling capability of Transformers, we convert the vertical EPI features ${\boldsymbol F}_{ver}$ to the sequences of ``tokens" for capturing spatial-angular correlation in $U$ and $H$ dimensions. 
As shown in Fig.~\ref{fig:Overview}(d), the vertical EPI features are passed through a linear projection matrix ${\it{\boldsymbol{W}}_{in}} \in \mathbb{R}^{\it{C} \times \it{D}}$, where $D$ denotes the embedding dimension of each token.
The projected EPI features are a sequence of tokens with a length of $UH$, i.e., ${\boldsymbol T}_{ver} \in \mathbb{R}^{UH \times D}$.
Following the PreNorm operation in \cite{PreNorm}, we also apply Layer Normalization (LN) before the attention calculation, and obtain the normalized tokens $\Bar{{\boldsymbol T}}_{ver} = {\rm LN}({\boldsymbol T}_{ver})$.

Afterwards, tokens ${\Bar{{\boldsymbol T}}}_{ver}$ are passed through the {\it Self-Attention} layer and transformed into the deep tokens involving non-local spatial-angular information along the vertical epipolar line. 
Specifically, ${\Bar{\boldsymbol T}}_{ver}$ need to be separately multiplied by ${\boldsymbol{W}_{Q}\in \mathbb{R}^{\it{D} \times \it{D}}}$, ${\boldsymbol{W}_{K}\in \mathbb{R}^{\it{D} \times \it{D}}}$ and ${\boldsymbol{W}_{V}\in \mathbb{R}^{\it{D} \times \it{D}}}$ to generate corresponding {\it{query}}, {\it{key}} and {\it{value}} components for self-attention calculation, i.e., ${{\boldsymbol Q}_{ver}={\Bar{{\boldsymbol T}}_{ver}}}{\it{{\boldsymbol W}}_{Q}}$, ${{\boldsymbol K}_{ver}={\Bar{{\boldsymbol T}}_{ver}}}{\it{{\boldsymbol W}}_{K}}$ and ${{\boldsymbol V}_{ver}={\Bar{{\boldsymbol T}}_{ver}}}{\it{{\boldsymbol W}}_{V}}$.

Given a {\it query} position $q=\{1,2,...,{\it U}{\it H}\}$ in ${\boldsymbol Q}_{ver}$ and a {\it key} position $k=\{1,2,...,{\it U}{\it H}\}$ in ${\boldsymbol K}_{ver}$, the corresponding response ${\boldsymbol A}_{ver}(q,k)\in \mathbb{R}$ measures the mutual similarity of the pairs by the dot-product operation, followed by a Softmax function to obtain the attention scores on the vertical EPI tokens. 
That is, 
\begin{equation}
\begin{split}
{{\boldsymbol A}_{ver}}(q,k) = 
{\rm Softmax
  ({\frac {\it{{\boldsymbol Q}_{ver}}(q)  \cdot \it{{\boldsymbol K}_{ver}}(k)^{{\rm T}} } 
    {\sqrt{ \it{D} }}})}.
\end{split}
\end{equation}

Based on the attention scores, the output of self-attention ${\boldsymbol T}'_{ver}$ can be calculated as the weighted sum of {\it{value}}. 
In summary, the calculation process of {\it Self-Attention} layer can be formulated as:
\begin{equation}\label{eq3}
\begin{split}
{{\boldsymbol T}'_{ver}} = {{\boldsymbol A}_{ver} \it{{\boldsymbol V}_{ver}}} + \it{{\boldsymbol T}_{ver}}.
\end{split}
\end{equation}

To further incorporate the spatial-angular correlation, following \cite{vaswani2017attention}, our {\it Basic-Transformer} unit also contains the multi-layer perception (MLP) and LN, and generates the enhanced $\hat{{\boldsymbol T}}_{ver}$ as:
\begin{equation}
\hat{\boldsymbol T}_{ver}={\rm MLP}({\rm LN}({\boldsymbol T}'_{ver}))+{\boldsymbol T}'_{ver}.
\end{equation}

At the end of the {\it Basic-Transformer} unit, the enhanced $\hat{\boldsymbol T}_{ver}$ is fed to another linear projection ${\boldsymbol W}_{out} \in \mathbb{R}^{\it{D} \times \it{C}}$, and reshaped into the size of ${UV \times H \times W \times C}$ for the subsequent {\it SpatialConv} layer.

\begin{figure}[t]
    \centering
    \includegraphics[width=8.0cm]{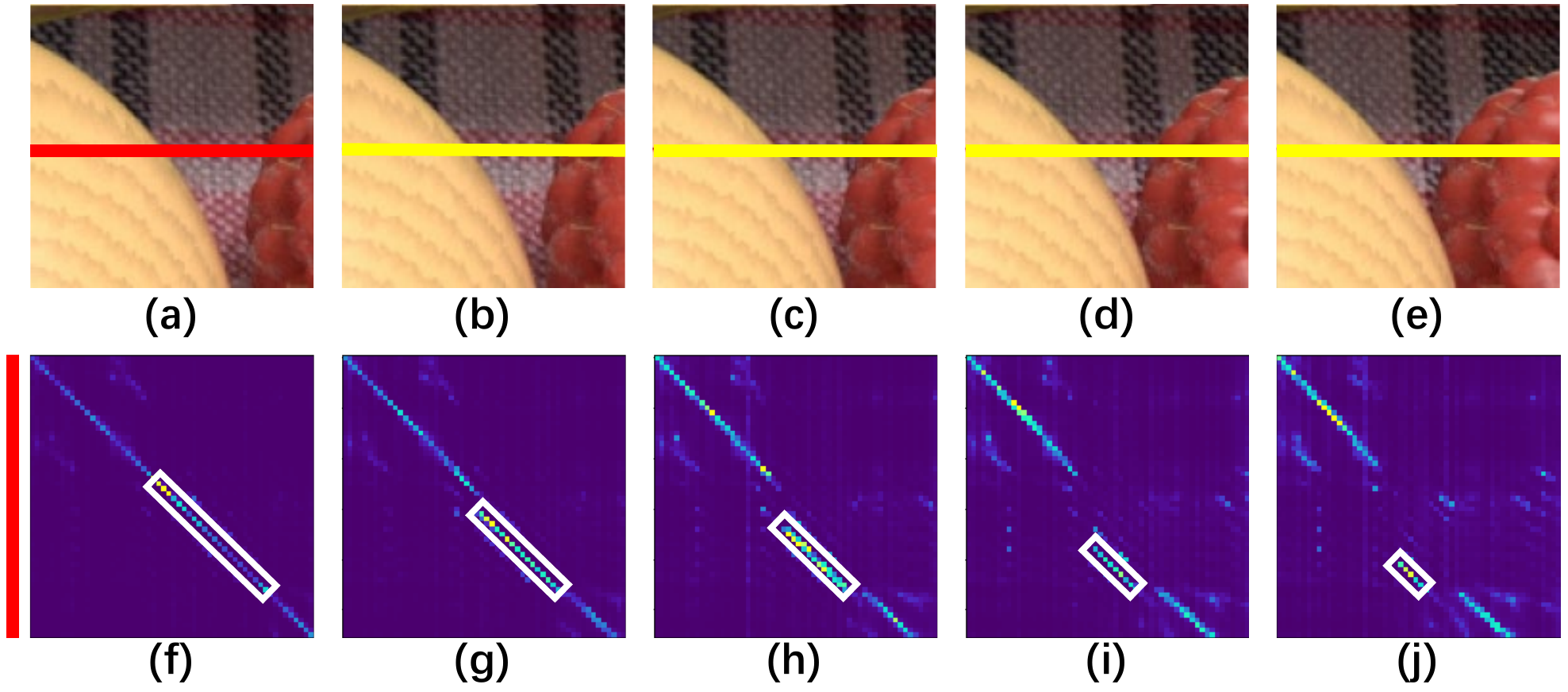}
    \caption{An example of the attention maps of a {\it Basic-Transformer} unit for the spatial-angular correlation. 
    Note that, the attention maps correspond to the correlation between the regions marked by the red and yellow strokes. }
    \label{fig:AttentionMap}
    \vspace{-0.2cm}
\end{figure}

\begin{table*}[t]
\caption{Quantitative comparison of different SR methods in terms of the number of parameters (\#Prm.) and PSNR/SSIM. Larger PSNR and SSIM values indicate higher SR quality. We mark the best results in \textcolor{red}{red} and the second best results in \textcolor{blue}{blue}.}
    \centering
    \scriptsize
    \renewcommand\arraystretch{1.22}
    \setlength{\tabcolsep}{1.2mm}{
    \begin{tabular}{|r|c|ccccc|ccccc|}
        \hline 
        \multirow{2}*{\textbf{ Methods~~~}} & {\textbf{\#Prm.(M)}} &  \multicolumn{5}{c|}{\scriptsize{\textbf{{2$\times$}}}} & \multicolumn{5}{c|}{\scriptsize{\textbf{{4$\times$}}}} \\
        \cline{3-12}
        & {\textbf{ 2$\times$/4$\times$}} & \textbf{EPFL}  & \textbf{HCInew}  & \textbf{HCIold}  & \textbf{INRIA} & \textbf{STFgantry} & \textbf{EPFL} & \textbf{HCInew} & \textbf{HCIold} & \textbf{INRIA} & \textbf{STFgantry} \\
        \hline
        \hline 
        \textit{Bicubic} & - / -
        & 29.74/.9376 & 31.89/.9356 & 37.69/.9785 & 31.33/.9577 & 31.06/.9498 
        & 25.14/.8324 & 27.61/.8517 & 32.42/.9344 & 26.82/.8867 & 25.93/.8452 \\
        \textit{VDSR} \cite{VDSR}  & 0.66 / 0.66
        & 32.50/.9598 & 34.37/.9561 & 40.61/.9867 & 34.43/,9741 & 35.54/.9789 
        & 27.25/.8777 & 29.31/.8823 & 34.81/.9515 & 29.19/.9204 & 28.51/.9009 \\
        \textit{EDSR} \cite{EDSR} & 38.6 / 38.9
        & 33.09/.9629 & 34.83/.9592 & 41.01/.9874 & 34.97/.9764 & 36.29/.9818 
        & 27.84/.8854 & 29.60/.8869 & 35.18/.9536 & 29.66/.9257 & 28.70/.9072\\
        \textit{RCAN} \cite{RCAN} & 15.3 / 15.4
        & 33.16/.9634 & 34.98/.9603 & 41.05/.9875 & 35.01/.9769 & 36.33/.9831
        & 27.88/.8863 & 29.63/.8886 & 35.20/.9548 & 29.76/.9276 & 28.90/.9131\\
        \hline
        \textit{resLF}\cite{resLF}  & 7.98 / 8.64
        & 33.62/.9706 & 36.69/.9739 & 43.42/.9932 & 35.39/.9804 & 38.36/.9904 
        & 28.27/.9035 & 30.73/.9107 & 36.71/.9682 & 30.34/.9412 & 30.19/.9372\\
        \textit{LFSSR} \cite{LFSSR} & 0.88 / 1.77
        & 33.68/.9744 & 36.81/.9749 & 43.81/.9938 & 35.28/.9832 & 37.95/.9898 
        & 28.27/.9118 & 30.72/.9145 & 36.70/.9696 & 30.31/.9467 & 30.15/.9426 \\
        \textit{LF-ATO} \cite{LFATO} & 1.22 / 1.36
        & 34.27/.9757 & 37.24/.9767 & 44.20/.9942 & 36.15/.9842 & 39.64/.9929 
        & 28.52/.9115 & 30.88/.9135 & 37.00/.9699 & 30.71/.9484 & 30.61/.9430\\
        \textit{{LF-InterNet} } \cite{LF-InterNet}  & 5.04 / 5.48
        & 34.14/.9760 & 37.28/.9763 & 44.45/.9946 & 35.80/.9843 & 38.72/.9909 
        & 28.67/.9162 & 30.98/.9161 & 37.11/.9716 & 30.64/.9491 & 30.53/.9409        \\
        \textit{LF-DFnet} \cite{LF-DFnet}  & 3.94 / 3.99
                    & 34.44/.9755 & 37.44/.9773 & 44.23/.9941 & 36.36/.9840 & 39.61/.9926 
                    & 28.77/.9165 & 31.23/.9196 & 37.32/.9718 & 30.83/.9503 & 31.15/.9494\\
        \textit{{MEG-Net}} \cite{MegNet} & 1.69 / 1.77
                    & 34.30/.9773 & 37.42/.9777 & 44.08/.9942 & 36.09/.9849 & 38.77/.9915
                    & 28.74/.9160 & 31.10/.9177 & 37.28/.9716 & 30.66/.9490 & 30.77/.9453 \\
        \textit{LF-IINet} \cite{LF-IINet} & 4.84 / 4.88
                     & 34.68/.9773 & 37.74/.9790 & 44.84/\textcolor{blue}{.9948} & 36.57/.9853 & 39.86/\textcolor{black}{.9936} 
                     & 29.11/.9188 & 31.36/.9208 & 37.62/.9734 & 31.08/.9515 & 31.21/.9502\\
        \textit{DPT} \cite{LF-DPT} & 3.73 / 3.78
                    & 34.48/.9758 & 37.35/.9771  & 44.31/.9943 & 36.40/.9843 & 39.52/.9926
                    & 28.93/.9170 & 31.19/.9188 & 37.39/.9721 & 30.96/.9503  & 31.14/.9488 \\

        \textit{LFT} \cite{LFT} & 1.11 / 1.16
                    & 34.80/\textcolor{blue}{{.9781}} 
					& 37.84/.9791
					& 44.52/\textcolor{black}{{.9945}} 
					& \textcolor{blue}{{36.59}}/\textcolor{blue}{{.9855}} 
					& \textcolor{blue}{{40.51}}/\textcolor{black}{{.9941}} 
					
					& {{29.25}}/\textcolor{blue}{{.9210}} 
					& \textcolor{blue}{{31.46}}/\textcolor{blue}{{.9218}} 
					& \textcolor{blue}{{37.63}}/\textcolor{blue}{{.9735}} 
					& {{31.20}}/{{.9524}} 
					& \textcolor{blue}{{31.86}}/\textcolor{blue}{{.9548}} \\
		\textit{{DistgSSR}} \cite{LF-Distg} & 3.53 / 3.58
		            & \textcolor{blue}{{34.81}}/\textcolor{red}{{.9787}}
		            & \textcolor{blue}{{37.96}}/\textcolor{blue}{{.9796}}
		            & \textcolor{blue}{44.94}/\textcolor{red}{.9949}
                    & \textcolor{blue}{36.59}/\textcolor{red}{.9859}
                    & 40.40/\textcolor{blue}{.9942}

                    & 28.99/\textcolor{black}{.9195}
                    & 31.38/.9217
                    & 37.56/.9732
                    & 30.99/.9519
                    & 31.65/.9535 \\

        \textit{LFSAV} \cite{LFSAV} & 1.22 / 1.54 
                    & 34.62/.9772 & 37.43/.9776 & 44.22/.9942 & 36.36/.9849 & 38.69/.9914 
                    
                    & \textcolor{red}{29.37}/\textcolor{red}{.9223} & 31.45/.9217 & 37.50/.9721 & \textcolor{blue}{31.27}/\textcolor{red}{.9531} & 31.36/.9505\\
		
		\textit{EPIT (ours)} & 1.42 / 1.47
                    & \textcolor{red}{{34.83}}/\textcolor{black}{{.9775}} 
					& \textcolor{red}{{38.23}}/\textcolor{red}{{.9810}} 
					& \textcolor{red}{{45.08}}/\textcolor{red}{{.9949}} 
					& \textcolor{red}{{36.67}}/\textcolor{black}{{.9853}} 
					& \textcolor{red}{{42.17}}/\textcolor{red}{{.9957}} 
					
					& \textcolor{blue}{{29.34}}/{{.9197}} 
					& \textcolor{red}{{31.51}}/\textcolor{red}{{.9231}} 
					& \textcolor{red}{{37.68}}/\textcolor{red}{{.9737}} 
					& \textcolor{red}{{31.37}}/\textcolor{blue}{{.9526}} 
					& \textcolor{red}{{32.18}}/\textcolor{red}{{.9571}} \\
					
		\hline
					
    \end{tabular}}
    \label{tab:spatialSR}
    \vspace{-0.4cm}
\end{table*}

\noindent{\bf Cross-View Similarity Analysis.}
Note that, the setting $\left[ {{\boldsymbol A}_{ver}}(q,1),...,{{\boldsymbol A}_{ver}}(q,{\it U}{\it H})  \right] \in \mathbb{R}^{1\times {\it U}{\it H}}$ represents the similarity scores of $q$ with all $k$ in $K_{ver}$, and thus can be re-organized as a slice of cross-view attention map according to the angular coordinate.
Inspired by this, we visualized the cross-view attention maps on an example scene in Fig.~\ref{fig:AttentionMap}.
The regions marked by the red stripe in Fig.~\ref{fig:AttentionMap}(a) are set as the {\it query} tokens, and the self-similarity (i.e., {\it key} are same as {\it query}) is ideally located at the diagonal, as shown in Fig.~\ref{fig:AttentionMap}(f). 
In contrast, the yellow stripes in Figs.~\ref{fig:AttentionMap}(b)-\ref{fig:AttentionMap}(e) are set as the {\it key} tokens, the corresponding cross-view similarities are shown in Figs.~\ref{fig:AttentionMap}(g)-\ref{fig:AttentionMap}(j). 
It can be observed that due to the foreground occlusions, the responses of the background appear as short lines (marked by the white boxes) parallel to the diagonal in each cross-view attention map, and both of the distance to the diagonal and the length of response regions adaptively change as the {\it key} view moves along the baseline, which demonstrates the disparity-awareness of our {\it Basic-Transformer} unit.

\subsubsection{Feature Upsampling}
Finally, we apply the pixel shuffling operation to increase the spatial resolution of LF features, and further employ a 3$\times$3 convolution to obtain the super-resolved LF image $\mathcal{L}_{SR}$. 
Following most existing works \cite{LF-InterNet, LF-Distg, LFT, LF-DFnet, LF-DPT, LF-IINet, MegNet, resLF, LFSSR}, we use the $L_1$ loss function to train our network due to its robustness to outliers \cite{anagun2019srlibrary}.

\begin{figure*}[t]
    \centering
    \includegraphics[width=17.4cm]{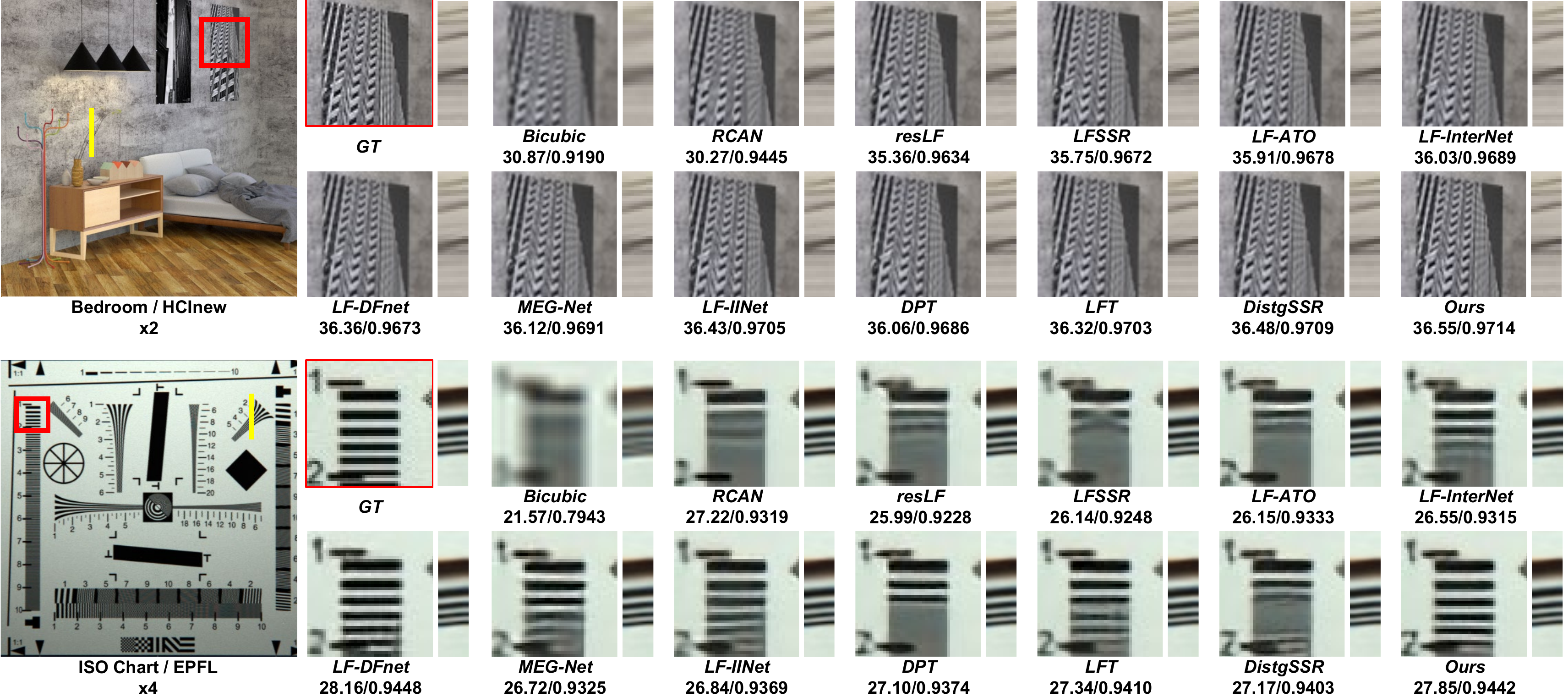}
    \caption{Qualitative comparison of different SR methods for 2$\times$/4$\times$ SR. }
    \vspace{-0.25cm}
    \label{fig:LF_visualx2x4}
\end{figure*}

\section{Experiments}

In this section, we first introduce the datasets and our implementation details, and then compare our method with state-of-the-art methods.
Next, we investigate the performance of different SR methods with respect to disparity variations.
Finally, we validate the effectiveness of our method through ablation studies.

\subsection{Datasets and Implementation Details}

We followed \cite{LF-DFnet, LF-Distg, LF-IINet, LF-DPT, LFT} to use five public LF datasets (EPFL \cite{EPFL}, HCInew \cite{HCInew}, HCIold \cite{HCIold}, INRIA \cite{INRIA}, STFgantry \cite{STFgantry}) in the experiments. 
All LFs in these datasets have an angular resolution of 9$\times$9. 
Unless specifically mentioned, we extracted the central 5$\times$5 SAIs for training and test. 
In the training stage, we cropped each SAI into patches of size 64$\times$64$/$128$\times$128, and performed 0.5$\times/$0.25$\times$ bicubic downsampling to generate the LR patches for 2$\times/$4$\times$ SR, respectively. 
We used peak signal-to-noise ratio (PSNR) and structural similarity (SSIM) \cite{SSIM} as quantitative metrics for performance evaluation.
To obtain the metric score for a dataset with $M$ scenes, we first calculated the metric of each scene by averaging the scores over all the SAIs separately, and then obtained the score for this dataset by averaging the scores over the $M$ scenes.

We adopted the same training settings for all experiments, i.e., Xavier initialization algorithm \cite{Xavier} and Adam optimizer \cite{Adam} with $\beta_{1}=0.9$, $\beta_{2}=0.999$. 
The initial learning rate was set to 2$\times 10^{-4}$ and decreased by a factor of 0.5 every 15 epochs. 
During the training phase, we performed random horizontal flipping, vertical flipping, and 90-degree rotation to augment the training data.
All models were implemented in the PyTorch framework and trained from scratch for 80 epochs with 2 Nvidia RTX 2080Ti GPUs.


\subsection{Comparisons on Benchmark Datasets}

We compare our method to 14 state-of-the-art methods, including 3 single image SR methods \cite{VDSR, EDSR, RCAN} and 11 LF image SR methods \cite{resLF, LFSSR, LFATO, LF-InterNet, LF-DFnet, MegNet, LF-IINet, LF-DPT, LFT, LF-Distg, LFSAV}.

\noindent{\bf Quantitative Results.}
A quantitative comparison among different methods is shown in Tabel~\ref{tab:spatialSR}. 
Our EPIT with a small model size (i.e., 1.42M$/$1.47M for 2$\times /$4$\times$ SR) achieves state-of-the-art PSNR and SSIM scores on almost all the datasets for both 2$\times$ and 4$\times$ SR. 
It is worth noting that LFs in the STFgantry dataset \cite{STFgantry} have larger disparity variations, and are thus more challenging.  
Our EPIT significantly outperforms all the compared methods and achieves 1.66dB$/$0.32dB PSNR improvements over the second top-performing method LFT for 2$\times/$4$\times$ SR, respectively, which demonstrates the powerful capacity of our EPIT in non-local correlation modeling.

\noindent{\bf Qualitative Results.}
Figure~\ref{fig:LF_visualx2x4} shows the qualitative results achieved by different methods for 2$\times$/4$\times$ SR. 
It can be observed from the zoom-in regions that single image SR method RCAN \cite{RCAN} cannot recover the textures and details in the SR images. 
In contrast, our EPIT can incorporate sub-pixel correspondence among SAIs and generate more faithful details with fewer artifacts. 
Compared to most LF image SR methods, our EPIT can generate superior visual results with high angular consistency. 
Please refer to the supplemental material for additional visual comparisons.

\begin{figure}[t]
    \centering
    \includegraphics[width=8.4cm]{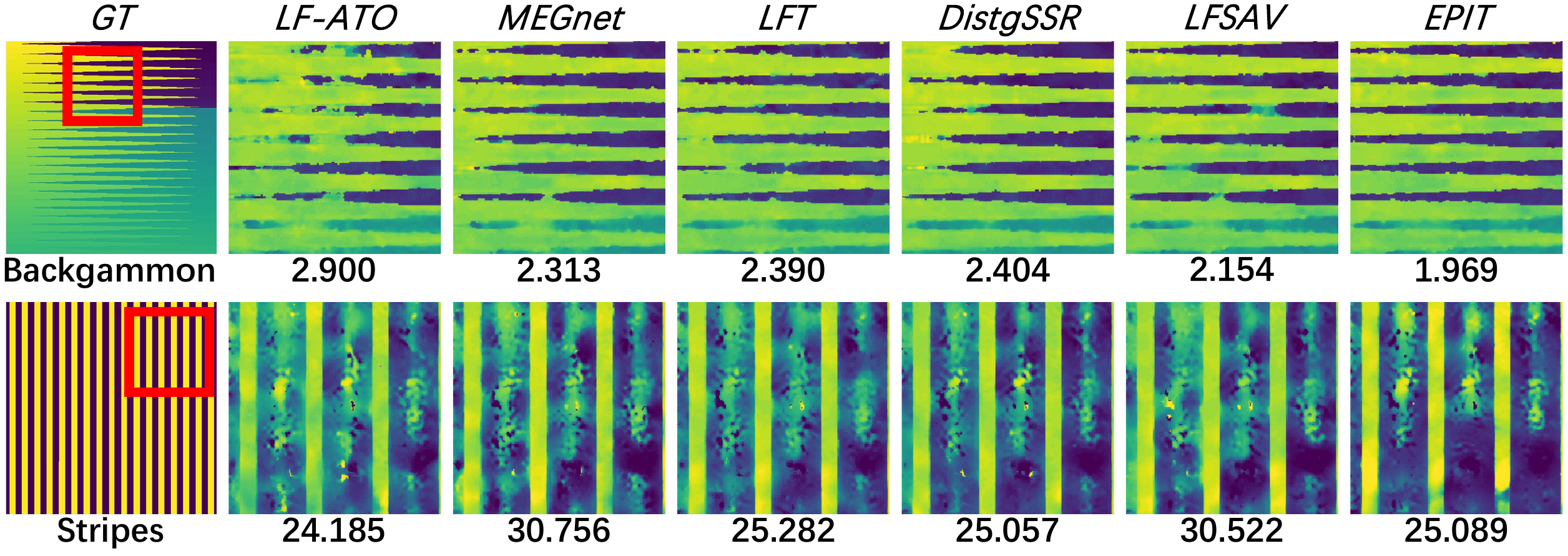}
    \caption{Quantitative and qualitative (MSE) comparisons of disparity estimation results achieved by SPO \cite{SPO} using different SR results. The MSE ($\downarrow$) is the mean square error multiplied by 100. 
    }
    \label{fig:viewconsistency}
\end{figure}

\begin{table}[]
    \centering
    \scriptsize
    \caption{PSNR values achieved by DistgSSR \cite{LF-Distg} and our EPIT with different angular resolution for 4$\times$ SR. }
    \setlength{\tabcolsep}{1.0mm}{
    \begin{tabular}{|c|cc|cc|cc|cc|cc|}
        \hline
        \multirow{2}*{Input} & \multicolumn{2}{c|}{{\bf EPFL}} & \multicolumn{2}{c|}{{\bf HCInew}} & \multicolumn{2}{c|}{{\bf HCIold}} & \multicolumn{2}{c|}{{\bf INRIA}} & \multicolumn{2}{c|}{{\bf STFgantry}}\\
        \cline{2-11}
         & \cite{LF-Distg} & Ours & \cite{LF-Distg} & Ours & \cite{LF-Distg} & Ours & \cite{LF-Distg} & Ours & \cite{LF-Distg} & Ours \\
        \hline
        2$\times$2 & 28.27 & {-0.05}  & 30.80 & {+0.04}     & 36.77 & {+0.17}     & 30.55 & -0.03     & 30.74 & +0.56  \\
        3$\times$3 & 28.67 & {+0.03}  & 31.07 & {+0.19}     & 37.18 & {+0.19}     & 30.83 & +0.11     & 31.12 & +0.74 \\
        4$\times$4 & 28.81 & {+0.23}  & 31.25 & {+0.15}     & 37.32 & {+0.20}     & 30.93 & +0.26     & 31.23 & +0.88 \\
        5$\times$5 & 28.99 & {+0.35}  & 31.38 & {+0.13}     & 37.56 & {+0.12}     & 30.99 & +0.38     & 31.65 & +0.56 \\
        6$\times$6 & 29.10 & {+0.33}  & 31.39 & {+0.18}     & 37.52 & {+0.26}     & 30.98 & +0.47     & 31.57 & +0.74 \\
        7$\times$7 & 29.38 & {+0.22}  & 31.43 & {+0.20}     & 37.65 & {+0.27}     & 31.18 & +0.33     & 31.63 & +0.77 \\
        8$\times$8 & 29.32 & {+0.28}  & 31.52 & {+0.14}     & 37.76 & {+0.24}     & 31.23 & +0.31     & 31.58 & +0.90 \\
        9$\times$9 & 29.41 & {+0.30}  & 31.48 & {+0.21}     & 37.80 & {+0.26}     & 31.22 & +0.34     & 31.66 & +0.84 \\
        \hline
    \end{tabular}}
    \label{tab:differentAng}
    \vspace{-0.3cm}
\end{table}

\begin{figure}[t]
    \centering
    \includegraphics[width=8.4cm]{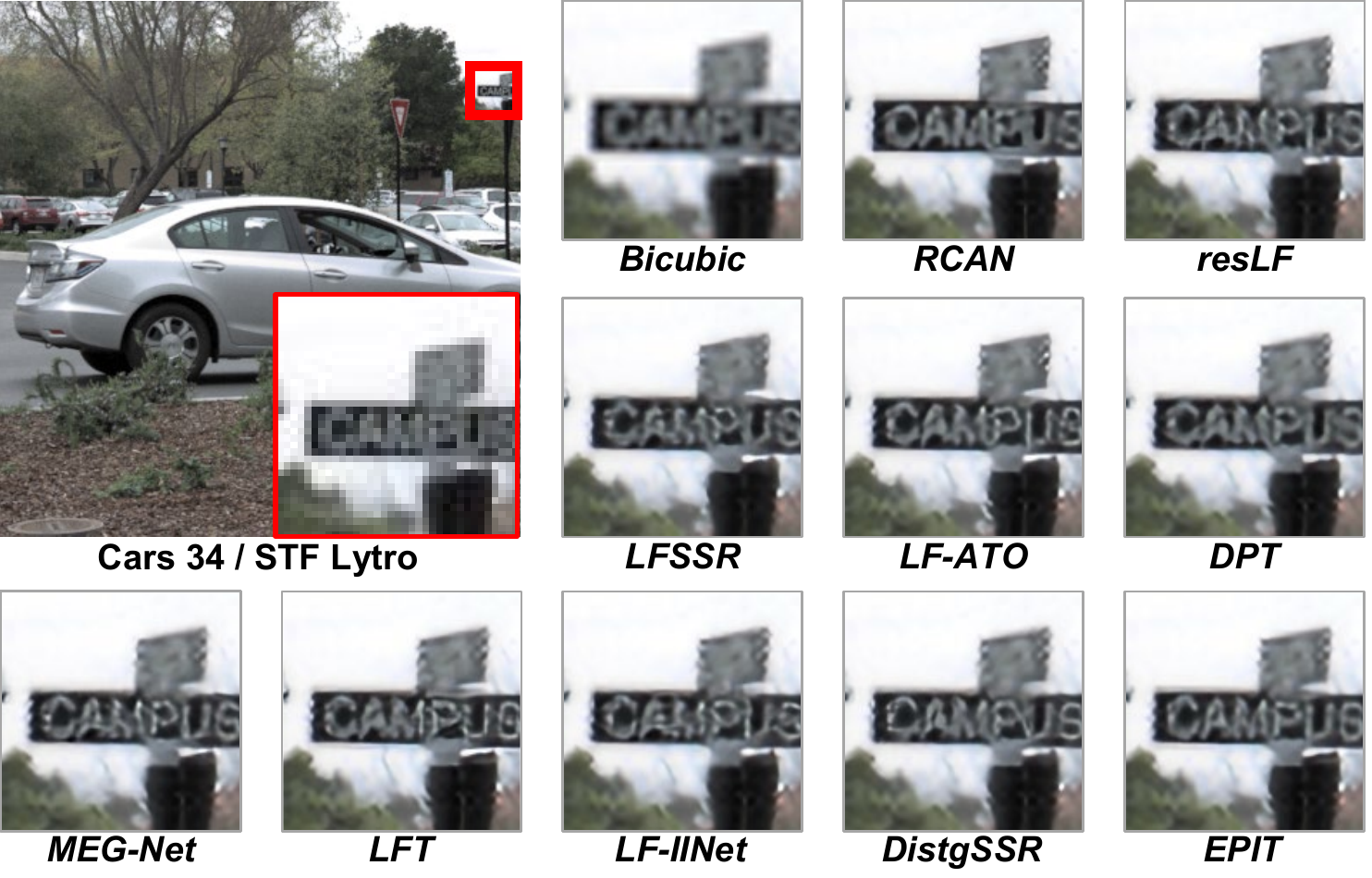}
    \caption{Visual comparison of different SR methods on real-world LF scenes for 4$\times$ SR. }
    \label{fig:RealVisualx4}
    \vspace{-0.4cm}
\end{figure}

\noindent{\bf Angular Consistency.}
We evaluate the angular consistency by using the 4$\times$ SR results on several challenging scenes (e.g., \texttt{Backgammon} and \texttt{Stripes}) in 4D LF benchmark \cite{HCInew} for disparity estimation. 
As shown in Fig.~\ref{fig:viewconsistency}, our EPIT achieves competitive MSE scores on these challenging scenes, which demonstrates the superiority of our EPIT on angular consistency.

\noindent{\bf Performance with Different Angular Resolution. }
Since the angular resolution of LR images can vary significantly with different LF devices, we compare our method to DistgSSR \cite{LF-Distg} on LFs with different angular resolutions. 
It can be observed from Table~\ref{tab:differentAng} that our method achieves higher PSNR values than DistgSSR on almost all the datasets with each angular resolution (except on the EPFL and INRIA datasets with 2$\times$2 input LFs). 
The consistent performance improvements demonstrate that our EPIT can well model the spatial-angular correlation with various angular resolutions.
More comparisons and discussions are provided in the supplemental material.

\begin{figure*}[t]
    \centering
    \includegraphics[width=17.4cm]{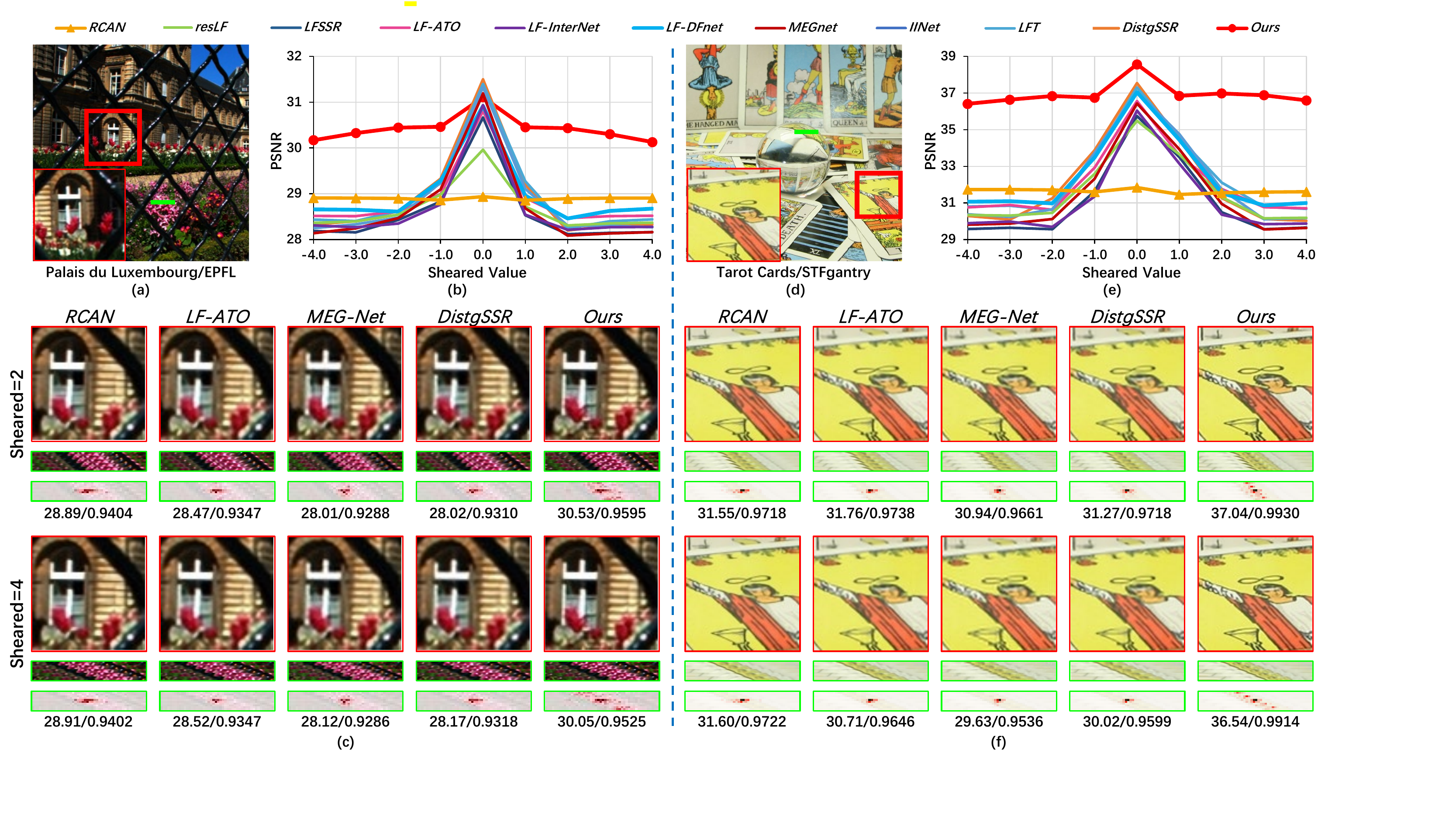}
    \caption{Performance comparison and local attribution maps of different SR methods on two representative scenes with different shearing values for 2$\times$ SR. 
    Here, we plot the performance curve to quantitatively measure the effect of disparity variations on LFs, and present the visual results and corresponding attribution maps under sheared value$=$2, 4. }
    \label{fig:LFSheared}
    \vspace{-0.2cm}
\end{figure*}

\noindent{\bf Performance on Real-World LF Scenes.}
We compare our method to state-of-the-art methods under real-world degradation by directly applying them to LFs in the STFlytro dataset \cite{raj2016stanford}. 
Since no groundtruth HR images are available in this dataset, we present the LR input and their super-resolved results in Fig.~\ref{fig:RealVisualx4}. 
It can be observed that our method can recover more faithful details and generate more clear letters than other methods. 
Since the LF structure keeps unchanged under both bicubic and real-world degradation, our method can learn the spatial-angular correlation from bicubicly downsampled training data, and well generalize to LF images under real degradation.

\subsection{Robustness to Large Disparity Variations} 
\label{sec:ShearedDisparity}

Considering the parallax structure of LF images, we followed the shearing operation in existing works \cite{ShearedEPI, wu2021revisiting} to linearly change the overall disparity range of LF datasets. 
Note that, the content of SAIs maintain unchanged after the shearing operation, and thus we can quantitatively investigate the performance of different SR methods with respect to the disparity variations.

\noindent{\bf Quantitative \& Qualitative Comparison.}
Figure~\ref{fig:LFSheared} shows the quantitative and qualitative results of different SR methods with respect to sheared values, from which we can observe that:
1) Except for the single image SR method RCAN, all LF image SR methods suffer a performance drop when the absolute sheared value of LF images increases. 
That is because, large sheared values can result in more significant misalignment among LF images, and introduce difficulties in complementary information incorporation; 
2) As the absolute sheared value increases, the performance of existing LF image SR methods is even inferior to RCAN. 
The possible reason is that, these methods do not make full use of local spatial information, but rather rely on local angular information from adjacent views.
When the sheared value exceeds their receptive fields, the large disparities can make the spatial-angular correlation non-local and thus introduce challenges in complementary information incorporation;
3) Our EPIT performs much more robust to disparity variations and achieves the highest PSNR scores under all sheared values. 
More quantitative comparisons on the whole datasets can be referred to the supplemental material.

\noindent{\bf LAM Visualization.}
We used Local Attribution Map (LAM) \cite{LAM} to visualize the input regions that contribute to the SR results of different methods.
As shown in Fig.~\ref{fig:LFSheared}, we first specify the center of green stripes in HR images as the target regions, and then re-organize the corresponding attribution maps on LR images into the EPI patterns. 
It can be observed that RCAN achieves a larger receptive field along the spatial dimension than other compared methods, which supports the results in Figs.~\ref{fig:LFSheared}(b) and ~\ref{fig:LFSheared}(e) that RCAN achieves a relatively stable SR performance with different sheared values. 
It is worth noting that our EPIT can automatically incorporate the most relevant information from different views, and can learn the non-local spatial-angular correlation regardless of disparity variations.

\begin{figure}[t]
    \centering
    \includegraphics[width=8.4cm]{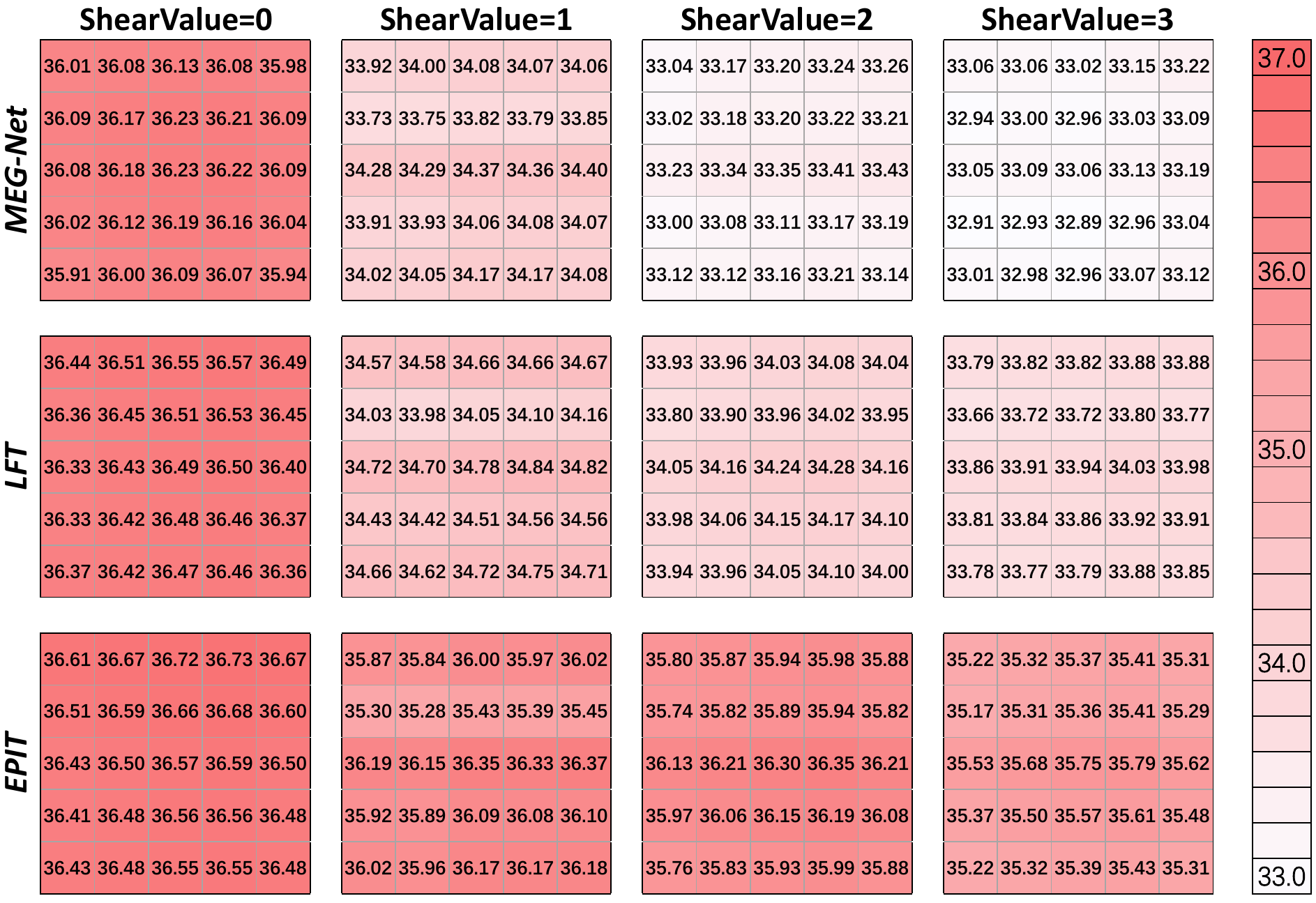}
    \caption{PSNR distribution among different SAIs achieved by MEG-Net \cite{MegNet}, LFT \cite{LFT}, and our EPIT on the INRIA dataset \cite{INRIA} for 2$\times$ SR. }
    \label{fig:perspective}
    \vspace{-0.2cm}
\end{figure}

\noindent{\bf Perspective Comparison.}
We compare the performance of MEG-Net, DistgSSR and our method with respect to different perspectives and sheared values (0 to 3).
It can be observed in Fig.~\ref{fig:perspective} that, both MEG-Net and DistgSSR suffer significant performance drops on all perspectives as the sheared value increases. 
In contrast, our EPIT can well handle the disparity variation problem, and achieve much higher PSNR values with a balanced distribution among different views regardless of the sheared values.




\subsection{Ablation Study}
\label{sec:ModelAnalysis}

In this subsection, we compare the performance of our EPIT with different variants to verify the effectiveness of our design choices, 
and additionally, investigate their robustness to large disparity variations.  

\noindent{\bf Horizontal/Vertical Basic-Transformer Units.}
We demonstrated the effectiveness of the horizontal and vertical {\it Basic-Transformer} units in our EPIT by separately removing them from our network.
Note that, without using horizontal or vertical {\it Basic-Transformer} unit, these variants cannot incorporate any information from the corresponding angular directions. 
As shown in Table~\ref{tab:ablation}, both variants {\it w/o-Horizontal} and {\it w/o-Vertical} suffer a decrease of 0.72dB in the INRIA dataset as compared to EPIT, which demonstrates the importance of exploiting spatial-angular correlations from all angular views.

\noindent{\bf Weight Sharing in Non-Local Cascading Blocks.}
We introduced the variant {\it w/o-Share} by removing the weight sharing between horizontal and vertical {\it Basic-Transformer} units.
As shown in Table~\ref{tab:ablation}, the additional parameters in variant {\it w/o-Share} do not introduce further performance improvement. 
It demonstrates that the weight sharing strategy between two directional {\it Basic-Transformer} units is beneficial and efficient to regularize the network.

\noindent{\bf SpatialConv in Non-Local Cascading Blocks.}
We introduced the variant {\it w/o-Local} by removing the {\it SpatialConv} layers from our EPIT, and we adjusted the channel number to make the model size of this variant not smaller than the main model.
As shown in Table~\ref{tab:ablation}, the {\it SpatialConv} has a significant influence on the SR performance, e.g., the variant {\it w/o-Local} suffers a 0.41dB PSNR drop on the EPFL dataset. 
It demonstrates that local context information is crucial to the SR performance, and the simple convolutions can fully incorporate the spatial information from each SAI.

\noindent{\bf Basic-Transformer in Non-Local Cascading Blocks.}
We introduced the variant {\it w/o-Trans} by replacing Basic-Transformer in Non-Local Blocks with cascaded convolutions. As shown in Table~\ref{tab:ablation}, {\it w/o-Trans} suffers a most significant performance drop as the sheared value increases, which demonstrates the effectiveness of the Basic-Transformer in incorporating global information on the EPIs. 

\noindent{\bf Basic-Transformer Number.}
We introduced the variants {\it with-n-Block} ({\it n}=1,2,3) by retaining {\it n} Non-Local Blocks. Results in Table~\ref{tab:ablation} show the effectiveness of our EPIT (having 5 Non-Local Blocks) with higher-order spatial-angular correlation modeling capability. 

\begin{table}[]
\caption{The PSNR scores achieved by different variants of our EPIT on the LFs with different shearing values for 2$\times$ SR. We adjusted the channel number of each variant to make its model size (i.e., \#Prm.) not smaller than EPIT for better validation.}
    \centering
    \scriptsize
    \renewcommand\arraystretch{1.2}
    \setlength{\tabcolsep}{1.4mm}{
    \begin{tabular}{|l|c|c|ccc|ccc|}
        \hline 
        \multirow{2}*{\bf Variants} & \multirow{2}*{\bf \#Prm.} & \multirow{2}*{\bf FLOPs} & \multicolumn{3}{c|}{{\bf EPFL (Sheared)}} & \multicolumn{3}{c|}{{\bf INRIA (Sheared)}}  \\ 
        \cline{4-9}
         &  & &  0 & 2 & 4 & 0 & 2 & 4\\
        \hline
        
        {\it w/o-Horiz}     & 1.42M & 80.20G    & 33.96 & 33.98 & 34.02   & 35.95 & 36.08 & 36.11 \\
        {\it w/o-Verti}     & 1.42M & 80.20G    & 34.01 & 33.94 & 33.87   & 35.95 & 35.97 & 36.02 \\
        {\it w/o-Share}     & 2.71M & 80.20G    & 34.80 & 34.63 & 34.51   & 36.66 & 36.72 & 36.45 \\
        {\it w/o-Local}     & 1.64M & 96.39G    & 34.42 & 34.36 & 34.27   & 36.36 & 36.40 & 36.25 \\
        \hline
        {\it w/o-Trans}     & 1.60M & 78.82G    & 33.90 & 31.32 & 31.74   & 35.95 & 33.28 & 33.55 \\
        {\it w-1-Block}     & 1.54M & 68.23G    & 33.97 & 34.24 & 34.08   & 35.84 & 36.19 & 35.93 \\
        {\it w-2-Block}     & 1.45M & 73.37G    & 34.19 & 34.36 & 34.29   & 35.98 & 36.27 & 35.99 \\
        {\it w-3-Block}     & 1.71M & 85.78G    & 34.64 & 34.51 & 34.45   & 36.53 & 36.47 & 36.22 \\
        \hline
        {{\it EPIT}}        & 1.42M & 74.96G   & \textcolor{black}{34.83} & \textcolor{black}{34.69} & \textcolor{black}{34.59} & \textcolor{black}{36.67} & \textcolor{black}{36.75} & \textcolor{black}{36.59} \\
         \hline
    \end{tabular}}
    \label{tab:ablation}
    \vspace{-0.2cm}
\end{table}


\section{Conclusion}

In this paper, we propose a Transformer-based network for LF image SR. By modeling the dependencies between each pair of pixels on EPIs, our method can learn the spatial-angular correlation while achieving a global receptive field along the epipolar line. Extensive experimental results demonstrated that our method can not only achieve state-of-the-art SR performance on benchmark datasets, but also perform robust to large disparity variations.


\noindent{\bf Acknowledgment:} 
This work was supported in part by the Foundation for Innovative Research Groups of the National Natural Science Foundation of China under Grant 61921001.

{\small
\bibliographystyle{ieee_fullname}
\bibliography{EPIT}

\begin{thebibliography}{10}\itemsep=-1pt

\bibitem{alain2018light}
Martin Alain and Aljosa Smolic.
\newblock Light field super-resolution via lfbm5d sparse coding.
\newblock In {\em 2018 25th IEEE international conference on image processing
  (ICIP)}, pages 2501--2505, 2018.

\bibitem{anagun2019srlibrary}
Yildiray Anagun, Sahin Isik, and Erol Seke.
\newblock Srlibrary: comparing different loss functions for super-resolution
  over various convolutional architectures.
\newblock {\em Journal of Visual Communication and Image Representation},
  61:178--187, 2019.

\bibitem{attal2022learning}
Benjamin Attal, Jia-Bin Huang, Michael Zollh{\"o}fer, Johannes Kopf, and
  Changil Kim.
\newblock Learning neural light fields with ray-space embedding.
\newblock In {\em IEEE Conference on Computer Vision and Pattern Recognition
  (CVPR)}, pages 19819--19829, 2022.

\bibitem{berman2016non}
Dana Berman, Shai Avidan, et~al.
\newblock Non-local image dehazing.
\newblock In {\em IEEE Conference on Computer Vision and Pattern Recognition
  (CVPR)}, pages 1674--1682, 2016.

\bibitem{NonLocalMeans}
Antoni Buades, Bartomeu Coll, and J-M Morel.
\newblock A non-local algorithm for image denoising.
\newblock In {\em IEEE Conference on Computer Vision and Pattern Recognition
  (CVPR)}, volume~2, pages 60--65, 2005.

\bibitem{cai2018ray}
Zewei Cai, Xiaoli Liu, Xiang Peng, and Bruce~Z Gao.
\newblock Ray calibration and phase mapping for structured-light-field 3d
  reconstruction.
\newblock {\em Optics Express}, 26(6):7598--7613, 2018.

\bibitem{IPT}
Hanting Chen, Yunhe Wang, Tianyu Guo, Chang Xu, Yiping Deng, Zhenhua Liu, Siwei
  Ma, Chunjing Xu, Chao Xu, and Wen Gao.
\newblock Pre-trained image processing transformer.
\newblock In {\em IEEE Conference on Computer Vision and Pattern Recognition
  (CVPR)}, pages 12299--12310, 2021.

\bibitem{chen2016attention}
Liang-Chieh Chen, Yi Yang, Jiang Wang, Wei Xu, and Alan~L Yuille.
\newblock Attention to scale: Scale-aware semantic image segmentation.
\newblock In {\em IEEE Conference on Computer Vision and Pattern Recognition
  (CVPR)}, pages 3640--3649, 2016.

\bibitem{LFSAV}
Zhen Cheng, Yutong Liu, and Zhiwei Xiong.
\newblock Spatial-angular versatile convolution for light field reconstruction.
\newblock {\em IEEE Transactions on Computational Imaging}, 8:1131--1144, 2022.

\bibitem{LFZSSR}
Zhen Cheng, Zhiwei Xiong, Chang Chen, Dong Liu, and Zheng-Jun Zha.
\newblock Light field super-resolution with zero-shot learning.
\newblock In {\em IEEE Conference on Computer Vision and Pattern Recognition
  (CVPR)}, pages 10010--10019, 2021.

\bibitem{choi2021neural}
Suyeon Choi, Manu Gopakumar, Yifan Peng, Jonghyun Kim, and Gordon Wetzstein.
\newblock Neural 3d holography: Learning accurate wave propagation models for
  3d holographic virtual and augmented reality displays.
\newblock {\em ACM Transactions on Graphics (TOG)}, 40(6):1--12, 2021.

\bibitem{BM3D}
Kostadin Dabov, Alessandro Foi, Vladimir Katkovnik, and Karen Egiazarian.
\newblock Image denoising by sparse 3-d transform-domain collaborative
  filtering.
\newblock {\em IEEE Transactions on image processing}, 16(8):2080--2095, 2007.

\bibitem{ViT}
Alexey Dosovitskiy, Lucas Beyer, Alexander Kolesnikov, Dirk Weissenborn,
  Xiaohua Zhai, Thomas Unterthiner, Mostafa Dehghani, Matthias Minderer, Georg
  Heigold, Sylvain Gelly, et~al.
\newblock An image is worth 16x16 words: Transformers for image recognition at
  scale.
\newblock {\em International Conference on Learning and Representation (ICLR)},
  2015.

\bibitem{freedman2011image}
Gilad Freedman and Raanan Fattal.
\newblock Image and video upscaling from local self-examples.
\newblock {\em ACM Transactions on Graphics (TOG)}, 30(2):1--11, 2011.

\bibitem{fu2019dual}
Jun Fu, Jing Liu, Haijie Tian, Yong Li, Yongjun Bao, Zhiwei Fang, and Hanqing
  Lu.
\newblock Dual attention network for scene segmentation.
\newblock In {\em IEEE Conference on Computer Vision and Pattern Recognition
  (CVPR)}, pages 3146--3154, 2019.

\bibitem{glasner2009super}
Daniel Glasner, Shai Bagon, and Michal Irani.
\newblock Super-resolution from a single image.
\newblock In {\em IEEE Conference on Computer Vision and Pattern Recognition
  (CVPR)}, pages 349--356, 2009.

\bibitem{Xavier}
Xavier Glorot and Yoshua Bengio.
\newblock Understanding the difficulty of training deep feedforward neural
  networks.
\newblock In {\em Proceedings of the International Conference on Artificial
  Intelligence and Statistics}, pages 249--256, 2010.

\bibitem{LAM}
Jinjin Gu and Chao Dong.
\newblock Interpreting super-resolution networks with local attribution maps.
\newblock In {\em IEEE Conference on Computer Vision and Pattern Recognition
  (CVPR)}, pages 9199--9208, 2021.

\bibitem{gu2014weighted}
Shuhang Gu, Lei Zhang, Wangmeng Zuo, and Xiangchu Feng.
\newblock Weighted nuclear norm minimization with application to image
  denoising.
\newblock In {\em IEEE Conference on Computer Vision and Pattern Recognition
  (CVPR)}, pages 2862--2869, 2014.

\bibitem{guo2021deep}
Mantang Guo, Junhui Hou, Jing Jin, Jie Chen, and Lap-Pui Chau.
\newblock Deep spatial-angular regularization for light field imaging,
  denoising, and super-resolution.
\newblock {\em IEEE Transactions on Pattern Analysis and Machine Intelligence},
  44(10):6094--6110, 2021.

\bibitem{HCInew}
Katrin Honauer, Ole Johannsen, Daniel Kondermann, and Bastian Goldluecke.
\newblock A dataset and evaluation methodology for depth estimation on 4d light
  fields.
\newblock In {\em Asian Conference on Computer Vision (ACCV)}, pages 19--34,
  2016.

\bibitem{SEnet}
Jie Hu, Li Shen, and Gang Sun.
\newblock Squeeze-and-excitation networks.
\newblock In {\em IEEE Conference on Computer Vision and Pattern Recognition
  (CVPR)}, pages 7132--7141, 2018.

\bibitem{huang2015single}
Jia-Bin Huang, Abhishek Singh, and Narendra Ahuja.
\newblock Single image super-resolution from transformed self-exemplars.
\newblock In {\em IEEE Conference on Computer Vision and Pattern Recognition
  (CVPR)}, pages 5197--5206, 2015.

\bibitem{CCnet}
Zilong Huang, Xinggang Wang, Lichao Huang, Chang Huang, Yunchao Wei, and Wenyu
  Liu.
\newblock Ccnet: Criss-cross attention for semantic segmentation.
\newblock In {\em IEEE International Conference on Computer Vision (ICCV)},
  pages 603--612, 2019.

\bibitem{jin2022occlusion}
Jing Jin and Junhui Hou.
\newblock Occlusion-aware unsupervised learning of depth from 4-d light fields.
\newblock {\em IEEE Transactions on Image Processing}, 31:2216--2228, 2022.

\bibitem{LFATO}
Jing Jin, Junhui Hou, Jie Chen, and Sam Kwong.
\newblock Light field spatial super-resolution via deep combinatorial geometry
  embedding and structural consistency regularization.
\newblock In {\em IEEE Conference on Computer Vision and Pattern Recognition
  (CVPR)}, pages 2260--2269, 2020.

\bibitem{FS_GAF}
Jing Jin, Junhui Hou, Jie Chen, Huanqiang Zeng, Sam Kwong, and Jingyi Yu.
\newblock Deep coarse-to-fine dense light field reconstruction with flexible
  sampling and geometry-aware fusion.
\newblock {\em IEEE Transactions on Pattern Analysis and Machine Intelligence},
  2020.

\bibitem{kalantari2016learning}
Nima~Khademi Kalantari, Ting-Chun Wang, and Ravi Ramamoorthi.
\newblock Learning-based view synthesis for light field cameras.
\newblock {\em ACM Transactions on Graphics (TOG)}, 35(6):1--10, 2016.

\bibitem{khan2021differentiable}
Numair Khan, Min~H Kim, and James Tompkin.
\newblock Differentiable diffusion for dense depth estimation from multi-view
  images.
\newblock In {\em IEEE Conference on Computer Vision and Pattern Recognition
  (CVPR)}, pages 8912--8921, 2021.

\bibitem{VDSR}
Jiwon Kim, JungKwon Lee, and KyoungMu Lee.
\newblock Accurate image super-resolution using very deep convolutional
  networks.
\newblock In {\em IEEE Conference on Computer Vision and Pattern Recognition
  (CVPR)}, pages 1646--1654, 2016.

\bibitem{Adam}
DiederikP Kingma and Jimmy Ba.
\newblock Adam: A method for stochastic optimization.
\newblock {\em International Conference on Learning and Representation (ICLR)},
  2015.

\bibitem{leistner2022towards}
Titus Leistner, Radek Mackowiak, Lynton Ardizzone, Ullrich K{\"o}the, and
  Carsten Rother.
\newblock Towards multimodal depth estimation from light fields.
\newblock In {\em IEEE Conference on Computer Vision and Pattern Recognition
  (CVPR)}, pages 12953--12961, 2022.

\bibitem{LF_two_plane}
Marc Levoy and Pat Hanrahan.
\newblock Light field rendering.
\newblock In {\em Proceedings of the 23rd annual conference on Computer
  graphics and interactive techniques}, pages 31--42, 1996.

\bibitem{liang2015light}
Chia-Kai Liang and Ravi Ramamoorthi.
\newblock A light transport framework for lenslet light field cameras.
\newblock {\em ACM Transactions on Graphics (TOG)}, 34(2):1--19, 2015.

\bibitem{SwinIR}
Jingyun Liang, Jiezhang Cao, Guolei Sun, Kai Zhang, Luc Van~Gool, and Radu
  Timofte.
\newblock Swinir: Image restoration using swin transformer.
\newblock In {\em IEEE International Conference on Computer Vision Workshops
  (ICCVW)}, pages 1833--1844, 2021.

\bibitem{LFT}
Zhengyu Liang, Yingqian Wang, Longguang Wang, Jungang Yang, and Shilin Zhou.
\newblock Light field image super-resolution with transformers.
\newblock {\em IEEE Signal Processing Letters}, 29:563--567, 2022.

\bibitem{EDSR}
Bee Lim, Sanghyun Son, Heewon Kim, Seungjun Nah, and KyoungMu Lee.
\newblock Enhanced deep residual networks for single image super-resolution.
\newblock In {\em IEEE Conference on Computer Vision and Pattern Recognition
  Workshops (CVPRW)}, pages 136--144, 2017.

\bibitem{LF-IINet}
Gaosheng Liu, Huanjing Yue, Jiamin Wu, and Jingyu Yang.
\newblock Intra-inter view interaction network for light field image
  super-resolution.
\newblock {\em IEEE Transactions on Multimedia}, pages 1--1, 2021.

\bibitem{SwinTransformer}
Ze Liu, Yutong Lin, Yue Cao, Han Hu, Yixuan Wei, Zheng Zhang, Stephen Lin, and
  Baining Guo.
\newblock Swin transformer: Hierarchical vision transformer using shifted
  windows.
\newblock In {\em IEEE International Conference on Computer Vision (ICCV)},
  pages 10012--10022, 2021.

\bibitem{liu2022video}
Ze Liu, Jia Ning, Yue Cao, Yixuan Wei, Zheng Zhang, Stephen Lin, and Han Hu.
\newblock Video swin transformer.
\newblock In {\em IEEE Conference on Computer Vision and Pattern Recognition
  (CVPR)}, pages 3202--3211, 2022.

\bibitem{LeakyReLU}
Andrew~L Maas, Awni~Y Hannun, Andrew~Y Ng, et~al.
\newblock Rectifier nonlinearities improve neural network acoustic models.
\newblock In {\em Proc. icml}, volume~30, page~3, 2013.

\bibitem{HDDRNet}
Nan Meng, HaydenKwokHay So, Xing Sun, and Edmund Lam.
\newblock High-dimensional dense residual convolutional neural network for
  light field reconstruction.
\newblock {\em IEEE Transactions on Pattern Analysis and Machine Intelligence},
  2019.

\bibitem{meng2020high}
Nan Meng, Xiaofei Wu, Jianzhuang Liu, and Edmund Lam.
\newblock High-order residual network for light field super-resolution.
\newblock In {\em Proceedings of the AAAI Conference on Artificial
  Intelligence}, volume~34, pages 11757--11764, 2020.

\bibitem{mitra2012light}
Kaushik Mitra and Ashok Veeraraghavan.
\newblock Light field denoising, light field superresolution and stereo camera
  based refocussing using a gmm light field patch prior.
\newblock In {\em 2012 IEEE Computer Society Conference on Computer Vision and
  Pattern Recognition Workshops}, pages 22--28, 2012.

\bibitem{naseer2021intriguing}
Muhammad~Muzammal Naseer, Kanchana Ranasinghe, Salman~H Khan, Munawar Hayat,
  Fahad Shahbaz~Khan, and Ming-Hsuan Yang.
\newblock Intriguing properties of vision transformers.
\newblock {\em Advances in Neural Information Processing Systems},
  34:23296--23308, 2021.

\bibitem{INRIA}
MikaelLe Pendu, Xiaoran Jiang, and Christine Guillemot.
\newblock Light field inpainting propagation via low rank matrix completion.
\newblock {\em IEEE Transactions on Image Processing}, 27(4):1981--1993, 2018.

\bibitem{raj2016stanford}
Abhilash~Sunder Raj, Michael Lowney, Raj Shah, and Gordon Wetzstein.
\newblock Stanford lytro light field archive, 2016.

\bibitem{EPFL}
Martin Rerabek and Touradj Ebrahimi.
\newblock New light field image dataset.
\newblock In {\em International Conference on Quality of Multimedia Experience
  (QoMEX)}, 2016.

\bibitem{rossi2018geometry}
Mattia Rossi and Pascal Frossard.
\newblock Geometry-consistent light field super-resolution via graph-based
  regularization.
\newblock {\em IEEE Transactions on Image Processing}, 27(9):4207--4218, 2018.

\bibitem{shi2022rethinking}
Shuwei Shi, Jinjin Gu, Liangbin Xie, Xintao Wang, Yujiu Yang, and Chao Dong.
\newblock Rethinking alignment in video super-resolution transformers.
\newblock {\em Advances in Neural Information Processing Systems}, 2022.

\bibitem{singh2014super}
Abhishek Singh, Fatih Porikli, and Narendra Ahuja.
\newblock Super-resolving noisy images.
\newblock In {\em IEEE Conference on Computer Vision and Pattern Recognition
  (CVPR)}, pages 2846--2853, 2014.

\bibitem{suhail2022light}
Mohammed Suhail, Carlos Esteves, Leonid Sigal, and Ameesh Makadia.
\newblock Light field neural rendering.
\newblock In {\em IEEE Conference on Computer Vision and Pattern Recognition
  (CVPR)}, pages 8269--8279, 2022.

\bibitem{STFgantry}
Vaibhav Vaish and Andrew Adams.
\newblock The (new) stanford light field archive.
\newblock {\em Computer Graphics Laboratory, Stanford University}, 6(7), 2008.

\bibitem{vaswani2017attention}
Ashish Vaswani, Noam Shazeer, Niki Parmar, Jakob Uszkoreit, Llion Jones,
  Aidan~N Gomez, {\L}ukasz Kaiser, and Illia Polosukhin.
\newblock Attention is all you need.
\newblock {\em Advances in Neural Information Processing Systems}, 30, 2017.

\bibitem{PAM}
Longguang Wang, Yulan Guo, Yingqian Wang, Zhengfa Liang, Zaiping Lin, Jungang
  Yang, and Wei An.
\newblock Parallax attention for unsupervised stereo correspondence learning.
\newblock {\em IEEE Transactions on Pattern Analysis and Machine Intelligence},
  2020.

\bibitem{PASSRnet}
Longguang Wang, Yingqian Wang, Zhengfa Liang, Zaiping Lin, Jungang Yang, Wei
  An, and Yulan Guo.
\newblock Learning parallax attention for stereo image super-resolution.
\newblock In {\em IEEE Conference on Computer Vision and Pattern Recognition
  (CVPR)}, 2019.

\bibitem{LF-DPT}
Shunzhou Wang, Tianfei Zhou, Yao Lu, and Huijun Di.
\newblock Detail preserving transformer for light field image super-resolution.
\newblock In {\em Proceedings of the AAAI Conference on Artificial
  Intelligence,}, 2022.

\bibitem{wang2018non}
Xiaolong Wang, Ross Girshick, Abhinav Gupta, and Kaiming He.
\newblock Non-local neural networks.
\newblock In {\em IEEE Conference on Computer Vision and Pattern Recognition
  (CVPR)}, pages 7794--7803, 2018.

\bibitem{LFNet}
Yunlong Wang, Fei Liu, Kunbo Zhang, Guangqi Hou, Zhenan Sun, and Tieniu Tan.
\newblock Lfnet: A novel bidirectional recurrent convolutional neural network
  for light-field image super-resolution.
\newblock {\em IEEE Transactions on Image Processing}, 27(9):4274--4286, 2018.

\bibitem{LF-Distg}
Yingqian Wang, Longguang Wang, Gaochang Wu, Jungang Yang, Wei An, Jingyi Yu,
  and Yulan Guo.
\newblock Disentangling light fields for super-resolution and disparity
  estimation.
\newblock {\em IEEE Transactions on Pattern Analysis and Machine Intelligence},
  2022.

\bibitem{LF-InterNet}
Yingqian Wang, Longguang Wang, Jungang Yang, Wei An, Jingyi Yu, and Yulan Guo.
\newblock Spatial-angular interaction for light field image super-resolution.
\newblock In {\em European Conference on Computer Vision (ECCV)}, pages
  290--308, 2020.

\bibitem{LF-DFnet}
Yingqian Wang, Jungang Yang, Longguang Wang, Xinyi Ying, Tianhao Wu, Wei An,
  and Yulan Guo.
\newblock Light field image super-resolution using deformable convolution.
\newblock {\em IEEE Transactions on Image Processing}, 30:1057--1071, 2020.

\bibitem{SSIM}
Zhou Wang, AlanC Bovik, HamidR Sheikh, and EeroP Simoncelli.
\newblock Image quality assessment: from error visibility to structural
  similarity.
\newblock {\em IEEE Transactions on Image Processing}, 13(4):600--612, 2004.

\bibitem{wanner2013variational}
Sven Wanner and Bastian Goldluecke.
\newblock Variational light field analysis for disparity estimation and
  super-resolution.
\newblock {\em IEEE Transactions on Pattern Analysis and Machine Intelligence},
  36(3):606--619, 2013.

\bibitem{HCIold}
Sven Wanner, Stephan Meister, and Bastian Goldluecke.
\newblock Datasets and benchmarks for densely sampled 4d light fields.
\newblock In {\em Vision, Modelling and Visualization (VMV)}, volume~13, pages
  225--226, 2013.

\bibitem{wizadwongsa2021nex}
Suttisak Wizadwongsa, Pakkapon Phongthawee, Jiraphon Yenphraphai, and Supasorn
  Suwajanakorn.
\newblock Nex: Real-time view synthesis with neural basis expansion.
\newblock In {\em IEEE Conference on Computer Vision and Pattern Recognition
  (CVPR)}, pages 8534--8543, 2021.

\bibitem{ShearedEPI}
Gaochang Wu, Yebin Liu, Qionghai Dai, and Tianyou Chai.
\newblock Learning sheared epi structure for light field reconstruction.
\newblock {\em IEEE Transactions on Image Processing}, 28(7):3261--3273, 2019.

\bibitem{wu2021revisiting}
Gaochang Wu, Yebin Liu, Lu Fang, and Tianyou Chai.
\newblock Revisiting light field rendering with deep anti-aliasing neural
  network.
\newblock {\em IEEE Transactions on Pattern Analysis and Machine Intelligence},
  2021.

\bibitem{SAAN}
Gaochang Wu, Yingqian Wang, Yebin Liu, Lu Fang, and Tianyou Chai.
\newblock Spatial-angular attention network for light field reconstruction.
\newblock {\em IEEE Transactions on Image Processing}, 30:8999--9013, 2021.

\bibitem{PreNorm}
Ruibin Xiong, Yunchang Yang, Di He, Kai Zheng, Shuxin Zheng, Chen Xing,
  Huishuai Zhang, Yanyan Lan, Liwei Wang, and Tieyan Liu.
\newblock On layer normalization in the transformer architecture.
\newblock In {\em International Conference on Machine Learning}, pages
  10524--10533, 2020.

\bibitem{yang2013fast}
Jianchao Yang, Zhe Lin, and Scott Cohen.
\newblock Fast image super-resolution based on in-place example regression.
\newblock In {\em IEEE Conference on Computer Vision and Pattern Recognition
  (CVPR)}, pages 1059--1066, 2013.

\bibitem{LFSSR}
HenryWingFung Yeung, Junhui Hou, Xiaoming Chen, Jie Chen, Zhibo Chen, and
  YukYing Chung.
\newblock Light field spatial super-resolution using deep efficient
  spatial-angular separable convolution.
\newblock {\em IEEE Transactions on Image Processing}, 28(5):2319--2330, 2018.

\bibitem{LFCNN}
Youngjin Yoon, HaeGon Jeon, Donggeun Yoo, JoonYoung Lee, and InSo Kweon.
\newblock Light-field image super-resolution using convolutional neural
  network.
\newblock {\em IEEE Signal Processing Letters}, 24(6):848--852, 2017.

\bibitem{yu2017light}
Jingyi Yu.
\newblock A light-field journey to virtual reality.
\newblock {\em IEEE MultiMedia}, 24(2):104--112, 2017.

\bibitem{MANA}
Jiyang Yu, Jingen Liu, Liefeng Bo, and Tao Mei.
\newblock Memory-augmented non-local attention for video super-resolution.
\newblock In {\em IEEE Conference on Computer Vision and Pattern Recognition
  (CVPR)}, pages 17834--17843, 2022.

\bibitem{zeyde2010single}
Roman Zeyde, Michael Elad, and Matan Protter.
\newblock On single image scale-up using sparse-representations.
\newblock In {\em International Conference on Curves and Surfaces}, pages
  711--730, 2010.

\bibitem{zhang2021learning}
Jingyang Zhang, Yao Yao, and Long Quan.
\newblock Learning signed distance field for multi-view surface reconstruction.
\newblock In {\em IEEE International Conference on Computer Vision (ICCV)},
  pages 6525--6534, 2021.

\bibitem{MegNet}
Shuo Zhang, Song Chang, and Youfang Lin.
\newblock End-to-end light field spatial super-resolution network using
  multiple epipolar geometry.
\newblock {\em IEEE Transactions on Image Processing}, 30:5956--5968, 2021.

\bibitem{resLF}
Shuo Zhang, Youfang Lin, and Hao Sheng.
\newblock Residual networks for light field image super-resolution.
\newblock In {\em IEEE Conference on Computer Vision and Pattern Recognition
  (CVPR)}, pages 11046--11055, 2019.

\bibitem{SPO}
Shuo Zhang, Hao Sheng, Chao Li, Jun Zhang, and Zhang Xiong.
\newblock Robust depth estimation for light field via spinning parallelogram
  operator.
\newblock {\em Computer Vision and Image Understanding}, 145:148--159, 2016.

\bibitem{RCAN}
Yulun Zhang, Kunpeng Li, Kai Li, Lichen Wang, Bineng Zhong, and Yun Fu.
\newblock Image super-resolution using very deep residual channel attention
  networks.
\newblock In {\em European Conference on Computer Vision (ECCV)}, pages
  286--301, 2018.

\bibitem{zhu2019revisiting}
Hao Zhu, Mantang Guo, Hongdong Li, Qing Wang, and Antonio Robles-Kelly.
\newblock Revisiting spatio-angular trade-off in light field cameras and
  extended applications in super-resolution.
\newblock {\em IEEE Transactions on Visualization and Computer Graphics},
  27(6):3019--3033, 2019.

\end{thebibliography}
}

{
\clearpage
\setcounter{section}{0}
\setcounter{figure}{1}
\setcounter{table}{0}

\renewcommand\thesection{\Alph{section}} 
\renewcommand\thetable{\Roman{table}}
\renewcommand\thefigure{\Roman{figure}}

\title{Learning Non-Local Spatial-Angular Correlation for Light Field \\Image Super-Resolution ({\textit{Supplemental Material}})}

\author{
}



\twocolumn[{%
	\renewcommand\twocolumn[1][]{#1}%
	\maketitle
	\vspace{-1.8cm}
    \begin{center}
		\centering
		\includegraphics[width=17.4cm]{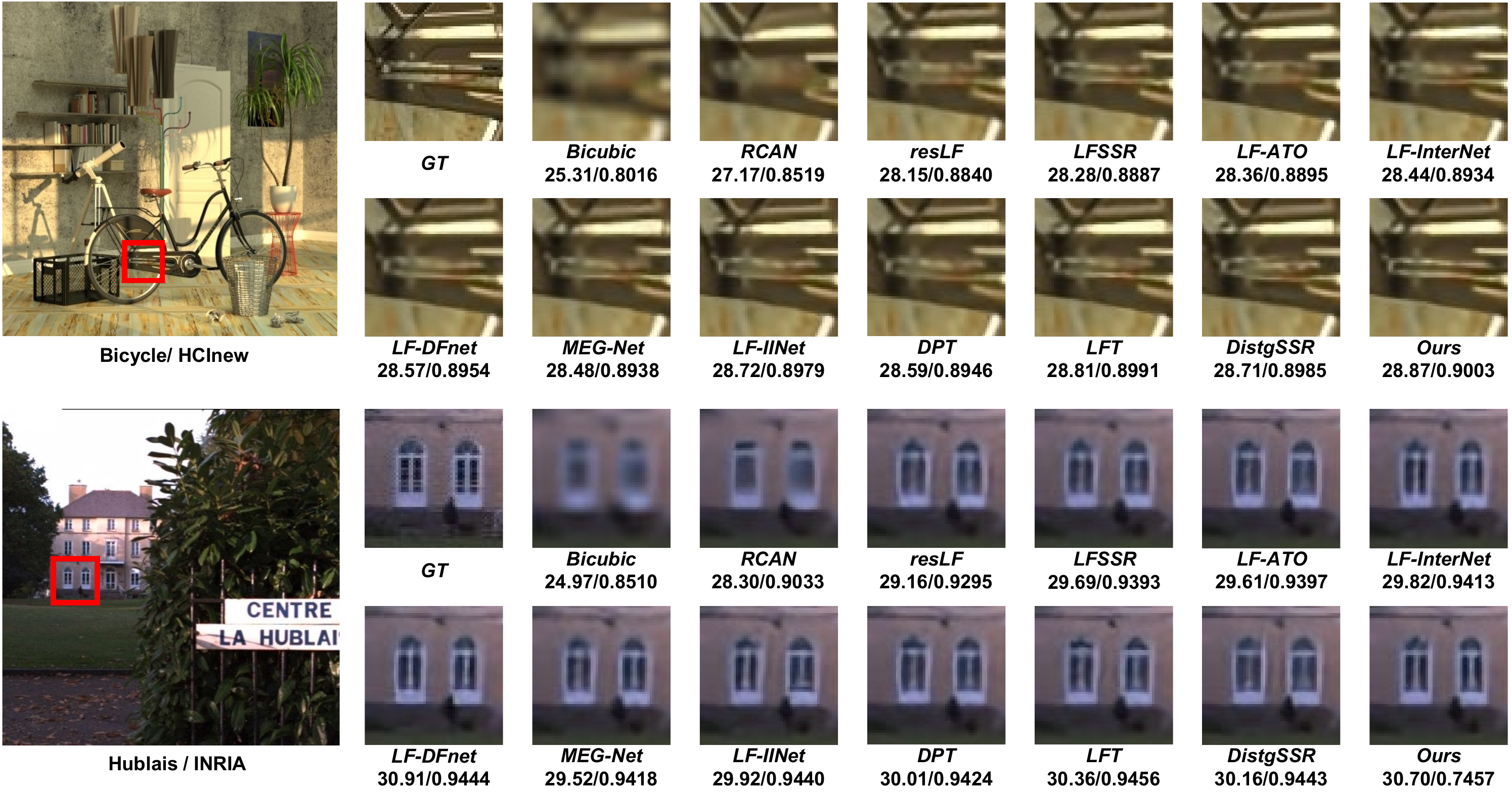}
        \vspace{-0.5cm}
		\label{fig:RealVisualx4_supp}
	\end{center}%
\centerline{Figure I. Qualitative comparison of different SR methods for 4$\times$ SR.}
\vspace{0.5cm}
}]

Section~\ref{sec:ComparisonBenchmarks} provides more visual comparisons on the light field (LF) datasets, and presents additional comparisons on LFs with different angular resolution. 
Section~\ref{sec:ComparsionSheared} presents detailed quantitative results of different methods on each dataset with various sheared values. 
Section~\ref{sec:LFASR} describes additional experiments for LF angular SR, and shows visual results achieved by different methods.

\section{Additional Comparisons on Benchmarks}
\label{sec:ComparisonBenchmarks}

\subsection{Qualitative Results}

In this subsection, we show more visual comparisons of 4$\times$ SR on the benchmark dataset in Fig. \textcolor{red}{I}.
It can be observed that the proposed EPIT recovers richer and more realistic details.

\subsection{Robustness to Different Angular Resolution}
\label{sec:DiffAng}

In the main body of our paper, we have illustrated that our EPIT (trained on central 5$\times$5 SAIs) achieves competitive PSNR scores on other angular resolutions, as compared to top-performing DistgSSR \cite{LF-Distg}. 
In Table~\ref{tab:methods_DiffAng}, we provide more quantitative results achieved by the state-of-the-art methods with different angular resolutions.

In addition, we train a series of EPIT models from scratch on 2$\times$2, 3$\times$3 and 4$\times$4 SAIs, respectively. 
It can be observed from Table~\ref{tab:differentAng_supp} that when using larger angular resolution SAIs as training data, e.g., 5$\times$5, our method can achieve better SR performance on different angular resolutions. 
That is because, more angular views are beneficial for our EPIT to learn the spatial-angular correlation better.
This phenomenon inspires us to explore the intrinsic mechanism of LF processing tasks in the future.

\begin{table}[]
    \centering
    \scriptsize
    \caption{PSNR/SSIM values achieved by different methods with different angular resolution for 4$\times$ SR. }
    \renewcommand\arraystretch{1.10}
    \setlength{\tabcolsep}{1.15mm}{
    \begin{tabular}{|cc|c|c|c|c|c|}
        \hline
         &  & \multicolumn{5}{c|}{\textbf{Methods}}\\
         \cline{3-7}
        \multicolumn{2}{|c|}{\multirow{-2}{*}{\textbf{Datasets}}}
        & {\it {resLF}} & {\it {LFSSR}} & {\it {MEG-Net}} & {\it {LFT}} & {\it {EPIT(ours)}} \\
        \hline
        \hline
        
        \multirow{8}*{\rotatebox{90}{\textbf{EPFL} \cite{EPFL}}}    
        & \multicolumn{1}{|c|}{2$\times$2}       & -            & 26.00/.8541  & 26.40/.8667 & 27.64/.8953 & 28.22/.9024 \\
        & \multicolumn{1}{|c|}{3$\times$3}       & 28.13/.9012  & 26.84/.8750  & 27.16/.8834 & 28.12/.9029 & 28.74/.9103 \\
        & \multicolumn{1}{|c|}{4$\times$4}       & -            & 27.62/.8930  & 28.04/.9036 & 28.43/.9087 & 29.04/.9164 \\
        & \multicolumn{1}{|c|}{5$\times$5}       & 28.27/.9035  & 28.27/.9118  & 28.74/.9160 & 29.85/.9210 & 29.34/.9197 \\
        & \multicolumn{1}{|c|}{6$\times$6}       & -            & 27.62/.8995  & 28.46/.9115 & 28.45/.9101 & 29.43/.9218 \\
        & \multicolumn{1}{|c|}{7$\times$7}       & 27.91/.9038  & 27.29/.8889  & 28.30/.9083 & 28.55/.9094 & 29.60/.9231 \\
        & \multicolumn{1}{|c|}{8$\times$8}       & -            & 27.06/.8834  & 28.15/.9061 & 28.37/.9064 & 29.60/.9240 \\
        & \multicolumn{1}{|c|}{9$\times$9}       & 26.07/.8881  & 26.95/.8810  & 28.12/.9046 & 28.45/.9071 & 29.71/.9246 \\
        \hline
        \hline
        
        \multirow{8}*{\rotatebox{90}{\textbf{HCInew} \cite{HCInew}}}
        & \multicolumn{1}{|c|}{2$\times$2}       & -            & 28.44/.8639  & 29.02/.8782 & 29.94/.8960 & 30.84/.9114 \\
        & \multicolumn{1}{|c|}{3$\times$3}       & 30.63/.9089  & 29.47/.8848  & 29.84/.8943 & 30.28/.9031 & 31.23/.9182 \\
        & \multicolumn{1}{|c|}{4$\times$4}       & -            & 30.22/.8997  & 30.68/.9094 & 30.51/.9065 & 31.40/.9213 \\
        & \multicolumn{1}{|c|}{5$\times$5}       & 30.73/.9107  & 30.72/.9145  & 31.10/.9177 & 31.46/.9218 & 31.51/.9231 \\
        & \multicolumn{1}{|c|}{6$\times$6}       & -            & 30.24/.9053  & 30.91/.9154 & 30.26/.9009 & 31.57/.9241 \\
        & \multicolumn{1}{|c|}{7$\times$7}       & 30.23/.9112  & 29.89/.8997  & 30.64/.9125 & 30.05/.8975 & 31.63/.9250 \\
        & \multicolumn{1}{|c|}{8$\times$8}       & -            & 29.68/.8969  & 30.48/.9105 & 29.81/.8923 & 31.66/.9256 \\
        & \multicolumn{1}{|c|}{9$\times$9}       & 27.84/.8967  & 29.46/.8942  & 30.34/.9087 & 29.77/.8916 & 31.69/.9260 \\
        \hline
        \hline                          
                                  
        \multirow{8}*{\rotatebox{90}{\textbf{HCIold} \cite{HCIold}}}
        & \multicolumn{1}{|c|}{2$\times$2}       & -            & 33.37/.9413  & 34.17/.9489 & 35.52/.9591 & 36.94/.9690  \\
        & \multicolumn{1}{|c|}{3$\times$3}       & 36.61/.9674  & 34.72/.9535  & 35.26/.9579 & 35.91/.9616 & 37.37/.9717 \\
        & \multicolumn{1}{|c|}{4$\times$4}       & -            & 35.80/.9615  & 36.42/.9662 & 36.15/.9634 & 37.52/.9729 \\
        & \multicolumn{1}{|c|}{5$\times$5}       & 36.71/.9682  & 36.70/.9696  & 37.28/.9716 & 37.63/.9735 & 37.68/.9737 \\
        & \multicolumn{1}{|c|}{6$\times$6}       & -            & 35.32/.9617  & 36.75/.9688 & 36.21/.9636 & 37.76/.9744  \\
        & \multicolumn{1}{|c|}{7$\times$7}       & 36.21/.968   & 34.94/.9578  & 36.35/.9662 & 36.10/.9629 & 37.92/.9749 \\
        & \multicolumn{1}{|c|}{8$\times$8}       & -            & 34.70/.9558  & 36.18/.9651 & 35.73/.9596 & 38.00/.9754 \\
        & \multicolumn{1}{|c|}{9$\times$9}       & 33.55/.9519  & 34.46/.9539  & 36.08/.9644 & 35.71/.9593 & 38.06/.9756 \\
        \hline
        \hline

        \multirow{8}*{\rotatebox{90}{\textbf{INRIA} \cite{INRIA}}}
        & \multicolumn{1}{|c|}{2$\times$2}       & -            & 27.83/.9035  & 28.31/.9125 & 29.99/.9378 & 30.52/.9418 \\
        & \multicolumn{1}{|c|}{3$\times$3}       & 30.33/.9413  & 28.78/.9201  & 29.16/.9264 & 30.35/.9424 & 30.94/.9472 \\
        & \multicolumn{1}{|c|}{4$\times$4}       & -            & 29.59/.9327  & 30.00/.9401 & 30.64/.9457 & 31.19/.9509 \\
        & \multicolumn{1}{|c|}{5$\times$5}       & 30.34/.9412  & 30.31/.9467  & 30.66/.9490 & 31.20/.9524 & 31.27/.9526 \\
        & \multicolumn{1}{|c|}{6$\times$6}       & -            & 29.50/.9356  & 30.38/.9443 & 30.61/.9457 & 31.45/.9533 \\
        & \multicolumn{1}{|c|}{7$\times$7}       & 29.82/.9398  & 29.05/.9269  & 30.13/.9415 & 30.56/.9443 & 31.51/.9539 \\
        & \multicolumn{1}{|c|}{8$\times$8}       & -            & 28.76/.9221  & 30.02/.9399 & 30.41/.9422 & 31.54/.9540 \\
        & \multicolumn{1}{|c|}{9$\times$9}       & 27.65/.9226  & 28.58/.9196  & 29.97/.9386 & 30.43/.9420 & 31.56/.9539 \\
        \hline
        \hline

        \multirow{8}*{\rotatebox{90}{\textbf{STFgantry} \cite{STFgantry}}}
        & \multicolumn{1}{|c|}{2$\times$2}       & -            & 27.29/.8710  & 28.15/.8944 & 29.69/.9263 & 31.30/.9468 \\
        & \multicolumn{1}{|c|}{3$\times$3}       & 30.05/.9348  & 28.81/.9064  & 29.22/.9161 & 30.05/.9316 & 31.86/.9534 \\
        & \multicolumn{1}{|c|}{4$\times$4}       & -            & 29.77/.9254  & 30.30/.9356 & 30.35/.9359 & 32.11/.9558 \\
        & \multicolumn{1}{|c|}{5$\times$5}       & 30.19/.9372  & 30.15/.9426  & 30.77/.9453 & 31.86/.9548 & 32.18/.9571 \\
        & \multicolumn{1}{|c|}{6$\times$6}       & -            & 29.79/.9320  & 30.58/.9428 & 30.01/.9289 & 32.31/.9580 \\
        & \multicolumn{1}{|c|}{7$\times$7}       & 29.71/.9375  & 29.40/.9257  & 30.25/.9393 & 29.53/.9208 & 32.40/.9585 \\
        & \multicolumn{1}{|c|}{8$\times$8}       & -            & 29.12/.9211  & 30.03/.9367 & 29.17/.9135 & 32.48/.9591 \\
        & \multicolumn{1}{|c|}{9$\times$9}       & 27.23/.9224  & 28.85/.9169  & 29.83/.9344 & 29.06/.9110 & 32.50/.9592 \\

        \hline
    \end{tabular}}
    \label{tab:methods_DiffAng}
\end{table}

\begin{table}[]
    \centering
    \begin{threeparttable}    
    \scriptsize
    \caption{PSNR/SSIM values achieved by our EPIT trained on LFs with different angular resolution for 4$\times$ SR. }
    \vspace{-0.16cm}
    \renewcommand\arraystretch{1.10}
    \setlength{\tabcolsep}{2.20mm}{
    \begin{tabular}{|cc|cccc|}
        \hline
        \multicolumn{2}{|c|}{\multirow{-0.5}{*}{\textbf{Datasets}}} & \multicolumn{4}{c|}{\it{EPIT(ours){*}}}\\
         \cline{3-6}
        
        & & 2$\times$2 & 3$\times$3 & 4$\times$4 & 5$\times$5\\
        \hline
        \hline

        \multirow{8}*{\rotatebox{90}{\textbf{EPFL} \cite{EPFL}}}
        & \multicolumn{1}{|c|}{2$\times$2}       & 28.40/.9037 & 28.45/.9040      & 28.33/.9034     & 28.22/.9024  \\
        & \multicolumn{1}{|c|}{3$\times$3}       & 28.61/.9076 & 28.75/.9090      & 28.67/.9090     & 28.74/.9103  \\
        & \multicolumn{1}{|c|}{4$\times$4}       & 28.69/.9108 & 28.90/.9131      & 28.86/.9137     & 29.04/.9164 \\
        & \multicolumn{1}{|c|}{5$\times$5}       & 28.81/.9124 & 29.08/.9152      & 29.06/.9162     & 29.34/.9197 \\
        & \multicolumn{1}{|c|}{6$\times$6}       & 28.81/.9133 & 29.13/.9168      & 29.12/.9180     & 29.43/.9218 \\
        & \multicolumn{1}{|c|}{7$\times$7}       & 28.88/.9137 & 29.24/.9176      & 29.24/.9190     & 29.60/.9231 \\
        & \multicolumn{1}{|c|}{8$\times$8}       & 28.86/.9140 & 29.25/.9184      & 29.25/.9198     & 29.60/.9240 \\
        & \multicolumn{1}{|c|}{9$\times$9}       & 28.92/.9141 & 29.32/.9188      & 29.34/.9204     & 29.71/.9246 \\
        \hline
        \hline

        \multirow{8}*{\rotatebox{90}{\textbf{HCInew} \cite{HCInew}}}
        & \multicolumn{1}{|c|}{2$\times$2}       & 30.81/.9109 & 30.86/.9116         & 30.86/.9116  & 30.84/.9114 \\
        & \multicolumn{1}{|c|}{3$\times$3}       & 30.84/.9124 & 31.06/.9157         & 31.09/.9162  & 31.23/.9182 \\
        & \multicolumn{1}{|c|}{4$\times$4}       & 30.86/.9132 & 31.14/.9174         & 31.21/.9184  & 31.40/.9213  \\
        & \multicolumn{1}{|c|}{5$\times$5}       & 30.86/.9134 & 31.19/.9184         & 31.27/.9197  & 31.51/.9231 \\
        & \multicolumn{1}{|c|}{6$\times$6}       & 30.86/.9134 & 31.21/.9190         & 31.32/.9205  & 31.57/.9241 \\
        & \multicolumn{1}{|c|}{7$\times$7}       & 30.85/.9133 & 31.23/.9194         & 31.35/.9211  & 31.63/.9250  \\
        & \multicolumn{1}{|c|}{8$\times$8}       & 30.86/.9133 & 31.24/.9197         & 31.37/.9215  & 31.66/.9256 \\
        & \multicolumn{1}{|c|}{9$\times$9}       & 30.85/.9132 & 31.25/.9199         & 31.39/.9219  & 31.69/.9260 \\
        \hline
        \hline

        \multirow{8}*{\rotatebox{90}{\textbf{HCIold} \cite{HCIold}}}
        & \multicolumn{1}{|c|}{2$\times$2}       & 36.83/.9683 & 36.85/.9682         & 36.81/.9679  & 36.94/.9690  \\
        & \multicolumn{1}{|c|}{3$\times$3}       & 36.92/.9688 & 37.13/.9701         & 37.14/.9702  & 37.37/.9717 \\
        & \multicolumn{1}{|c|}{4$\times$4}       & 36.95/.9692 & 37.21/.9708         & 37.27/.9712  & 37.52/.9729 \\
        & \multicolumn{1}{|c|}{5$\times$5}       & 37.01/.9695 & 37.31/.9714         & 37.39/.9718  & 37.68/.9737  \\
        & \multicolumn{1}{|c|}{6$\times$6}       & 37.00/.9696 & 37.33/.9717         & 37.44/.9723  & 37.76/.9744  \\
        & \multicolumn{1}{|c|}{7$\times$7}       & 37.00/.9696 & 37.40/.9719         & 37.52/.9726  & 37.92/.9749 \\
        & \multicolumn{1}{|c|}{8$\times$8}       & 36.99/.9696 & 37.41/.9721         & 37.56/.9729  & 38.00/.9754 \\
        & \multicolumn{1}{|c|}{9$\times$9}       & 36.99/.9697 & 37.44/.9722         & 37.60/.9730  & 38.06/.9756 \\
        \hline
        \hline

        \multirow{8}*{\rotatebox{90}{\textbf{INRIA} \cite{INRIA}}}
        & \multicolumn{1}{|c|}{2$\times$2}       & 30.63/.9429 & 30.66/.9431         & 30.58/.9427  & 30.52/.9418 \\
        & \multicolumn{1}{|c|}{3$\times$3}       & 30.82/.9458 & 30.91/.9465         & 30.87/.9466  & 30.94/.9472 \\
        & \multicolumn{1}{|c|}{4$\times$4}       & 30.90/.9472 & 31.04/.9484         & 31.02/.9489  & 31.19/.9509 \\
        & \multicolumn{1}{|c|}{5$\times$5}       & 30.95/.9483 & 31.14/.9498         & 31.14/.9506  & 31.27/.9526 \\
        & \multicolumn{1}{|c|}{6$\times$6}       & 30.94/.9484 & 31.17/.9503         & 31.18/.9511  & 31.45/.9533 \\
        & \multicolumn{1}{|c|}{7$\times$7}       & 30.93/.9485 & 31.20/.9506         & 31.22/.9515  & 31.51/.9539 \\
        & \multicolumn{1}{|c|}{8$\times$8}       & 30.92/.9484 & 31.22/.9507         & 31.24/.9517  & 31.54/.9540 \\
        & \multicolumn{1}{|c|}{9$\times$9}       & 30.91/.9481 & 31.22/.9506         & 31.26/.9516  & 31.56/.9539 \\
        \hline
        \hline

        \multirow{8}*{\rotatebox{90}{\textbf{STFgantry} \cite{STFgantry}}}
        & \multicolumn{1}{|c|}{2$\times$2}       & 30.84/.9432 & 31.03/.9449         & 31.09/.9452  & 31.30/.9468 \\
        & \multicolumn{1}{|c|}{3$\times$3}       & 30.93/.9447 & 31.39/.9493         & 31.49/.9503  & 31.86/.9534 \\
        & \multicolumn{1}{|c|}{4$\times$4}       & 31.02/.9459 & 31.56/.9510         & 31.69/.9523  & 32.11/.9558 \\
        & \multicolumn{1}{|c|}{5$\times$5}       & 30.99/.9459 & 31.58/.9518         & 31.74/.9534  & 32.18/.9571 \\
        & \multicolumn{1}{|c|}{6$\times$6}       & 31.03/.9460 & 31.68/.9525         & 31.85/.9541  & 32.31/.9580 \\
        & \multicolumn{1}{|c|}{7$\times$7}       & 31.03/.9459 & 31.70/.9526         & 31.90/.9545  & 32.40/.9585 \\
        & \multicolumn{1}{|c|}{8$\times$8}       & 31.04/.9459 & 31.73/.9528         & 31.96/.9549  & 32.48/.9591 \\
        & \multicolumn{1}{|c|}{9$\times$9}       & 31.02/.9457 & 31.74/.9529         & 31.97/.9550  & 32.50/.9592\\

        \hline
    \end{tabular}}
    \label{tab:differentAng_supp}
    \begin{tablenotes}[para,flushleft]  
        \item {*} Note that, ``{\it A}$\times${\it A}'' below ``EPIT(ours)'' denotes the models are trained on the LFs with corresponding angular resolution.
     \end{tablenotes} 
\end{threeparttable} 
\end{table}

\section{Additional Quantitative Comparison on Disparity Variations}
\label{sec:ComparsionSheared}

We have presented the performance comparison on two selected scenes with different shearing values for 2$\times$ SR in the main paper. 
Here, we provide quantitative results on each dataset in Table~\ref{tab:SheardDisparity} and Fig.~\ref{fig:shearedDisparity_x2}. 
It can be observed that our EPIT achieves more consistent performance than existing methods with respect to disparity variations on various datasets. 

\begin{table*}[]
    \centering
    \scriptsize
    \caption{Quantitative comparison of different SR methods on five datasets with different shearing values for 2$\times$ SR. We mark the best results in \textcolor{red}{red} and the second results in \textcolor{blue}{blue}.}
    \renewcommand\arraystretch{1.20}
    \setlength{\tabcolsep}{0.9mm}{
    \begin{tabular}{|cc|cccccccccccc|}
        \hline
         &  & \multicolumn{12}{c|}{\textbf{Methods}}\\
         \cline{3-14}
        \multicolumn{2}{|c|}{\multirow{-2}{*}{\textbf{Datasets}}} & 
        {\it {Bicubic}} & {\it {RCAN}} & {\it {resLF}} & {\it {LFSSR}} & {\it {LF-ATO}} & {\it {LF-InterNet}} & {\it {LF-DFnet}} 
        & {\it {MEG-Net}} & {\it {LF-IINet}} & {\it {LFT}} & {\it {DistgSSR}} & {\it {Ours}}\\
        
        \hline
        \hline
        \multirow{8}*{\rotatebox{90}{\textbf{EPFL} \cite{EPFL}}}
        &  \multicolumn{1}{|c|}{-4} 
        & 29.95/.9372 & \textcolor{blue}{33.47}/\textcolor{blue}{.9640} & 32.41/.9582 & 31.90/.9550 & 32.59/.9593 & 32.15/.9573 & 32.69/.9597 & 32.07/.9560 & 32.24 /.9579 & 32.48/.9587 & 32.29/.9583 & \textcolor{red}{34.52}/\textcolor{red}{.9734} \\
        &  \multicolumn{1}{|c|}{-3} 
        & 29.92/.9369 & \textcolor{blue}{33.45}/\textcolor{blue}{.9637} & 32.38/.9578 & 31.85/.9548 & 32.58/.9592 & 32.14/.9572 & 32.68/.9597 & 32.16/.9564 & 32.27/.9577 & 32.49/.9587 & 32.29/.9578 & \textcolor{red}{34.67}/\textcolor{red}{.9746} \\
        &  \multicolumn{1}{|c|}{-2} 
        & 29.89/.9369 & \textcolor{blue}{33.31}/\textcolor{blue}{.9632} & 32.36/.9587 & 31.92/.9561 & 32.37/.9589 & 32.06/.9571 & 32.47/.9592 & 32.17/.9574 & 32.37/.9589 & 32.35/.9587 & 32.65/.9618 & \textcolor{red}{34.64}/\textcolor{red}{.9749} \\
        &  \multicolumn{1}{|c|}{-1} 
        & 29.83/.9373 & 33.30/.9634 & 33.01/.9652 & 32.69/.9640 & 33.06/.9659 & 32.62/.9636 & \textcolor{blue}{33.41}/.9673 & 32.82/.9653 & 33.29/.9676 & 33.33/.9676 & 33.37/\textcolor{blue}{.9687} & \textcolor{red}{34.71}/\textcolor{red}{.9756} \\
        &  \multicolumn{1}{|c|}{0} 
        & 29.74/.9376 & 33.16/.9634 & 33.62/.9706 & 33.68/.9744 
        & 34.27/.9757 & 34.14/.9760 & 34.40/.9755 
        & 34.30/.9773 
        & 34.68/.9773 
        & 34.80/\textcolor{blue}{.9781} 
        & \textcolor{red}{34.81}/\textcolor{red}{.9787} 
        & \textcolor{red}{34.83}/.9775 \\
        &  \multicolumn{1}{|c|}{1} 
        & 29.87/.9373 & 33.16/.9629 & 32.81/.9644 & 32.70/.9639 & 32.67/.9656 & 32.57/.9642 & \textcolor{blue}{33.19}/.9669 & 32.76/.9647 & 33.12/.9663 & 33.18/.9675 & 33.01/\textcolor{blue}{.9681} & \textcolor{red}{34.66}/\textcolor{red}{.9760} \\
        &  \multicolumn{1}{|c|}{2} 
        & 29.91/.9370 & \textcolor{blue}{33.37}/\textcolor{blue}{.9633} & 32.28/.9579 & 31.87/.9548 & 32.47/.9597 & 32.00/.9569 & 32.45/.9593 & 31.85/.9560 & 32.15/.9577 & 32.42/.9598 & 32.04/.9581 & \textcolor{red}{34.69}/\textcolor{red}{.9750} \\
        &  \multicolumn{1}{|c|}{3} 
        & 29.94/.9370 & \textcolor{blue}{33.48}/\textcolor{blue}{.9638} & 32.32/.9575 & 31.85/.9543 & 32.56/.9591 & 32.09/.9569 & 32.61/.9594 & 31.84/.9545 & 32.19/.9574 & 32.43/.9585 & 32.17/.9578 & \textcolor{red}{34.73}/\textcolor{red}{.9747} \\
        &  \multicolumn{1}{|c|}{4} 
        & 29.98/.9372 & \textcolor{blue}{33.52}/\textcolor{blue}{.9641} & 32.40/.9579 & 31.97/.9550 & 32.57/.9592 & 32.15/.9572 & 32.68/.9597 & 31.93/.9554 & 32.19/.9575 & 32.46/.9586 & 32.15/.9579 & \textcolor{red}{34.59}/\textcolor{red}{.9736} \\
        
        \hline
        \hline
        \multirow{8}*{\rotatebox{90}{\textbf{HCInew} \cite{HCInew}}}
        &  \multicolumn{1}{|c|}{-4} & 30.83/.9343 & \textcolor{blue}{34.59}/\textcolor{blue}{.9611} & 33.34/.9533 & 32.57/.9494 & 33.37/.9545 & 32.99/.9525 & 33.62/.9554 & 32.91/.9510 & 33.34/.9534 & 33.35/.9541 & 33.03/.9523 & \textcolor{red}{36.77}/\textcolor{red}{.9782} \\
        &  \multicolumn{1}{|c|}{-3} & 30.81/.9342 & \textcolor{blue}{34.65}/\textcolor{blue}{.9609} & 33.45/.9543 & 32.61/.9501 & 33.58/.9558 & 33.06/.9523 & 33.75/.9562 & 33.16/.9527 & 33.44/.9542 & 33.51/.9551 & 33.43/.9554 & \textcolor{red}{37.05}/\textcolor{red}{.9791} \\
        &  \multicolumn{1}{|c|}{-2} & 30.83/.9344 & \textcolor{blue}{34.60}/.9605 & 33.50/.9594 & 32.58/.9548 & 33.13/.9599 & 32.91/.9563 & 33.41/.9609 & 33.33/.9588 & 33.80/.9618 & 33.37/.9609 & 33.76/\textcolor{blue}{.9644} & \textcolor{red}{36.98}/\textcolor{red}{.9792} \\
        &  \multicolumn{1}{|c|}{-1} & 30.74/.9349 & 34.42/.9603 & 35.00/.9704 & 34.19/.9691 & 34.87/.9716 & 34.29/.9690 & 35.59/.9739 & 34.51/.9716 & \textcolor{blue}{35.70}/.9748 & 35.49/.9747 & 35.68/\textcolor{blue}{.9754} & \textcolor{red}{37.21}/\textcolor{red}{.9815} \\
         &  \multicolumn{1}{|c|}{0} 
         & 31.89/.9356 & 34.98/.9603 & 36.69/.9739 & 36.81/.9749 
         & 37.24/.9767 & 37.28/.9763 & 37.44/.9773 
         & 37.42/.9777 
         & 37.74/.9790 
         & 37.84/.9791 
         & \textcolor{blue}{37.96}/\textcolor{blue}{.9796} 
         & \textcolor{red}{38.23}/\textcolor{red}{.9810} \\
         &  \multicolumn{1}{|c|}{1} & 30.73/.9350 & 34.14/.9602 & 34.04/.9649 & 33.90/.9639 & 33.41/.9660 & 33.63/.9633 & 34.30/.9681 & 34.06/.9659 & \textcolor{blue}{34.64}/.9682 & 34.33/\textcolor{blue}{.9694} & 34.30/.9691 & \textcolor{red}{36.83}/\textcolor{red}{.9792} \\
         &  \multicolumn{1}{|c|}{2} & 30.79/.9344 & \textcolor{blue}{34.30}/\textcolor{blue}{.9605} & 32.99/.9547 & 32.64/.9509 & 32.84/.9566 & 32.65/.9527 & 32.80/.9560 & 32.43/.9517 & 32.99/.9546 & 33.10/.9571 & 32.31/.9546 & \textcolor{red}{36.31}/\textcolor{red}{.9787} \\
         &  \multicolumn{1}{|c|}{3} & 30.77/.9341 & \textcolor{blue}{34.39}/\textcolor{blue}{.9609} & 33.17/.9523 & 32.70/.9493 & 33.32/.9545 & 33.03/.9523 & 33.51/.9553 & 32.59/.9492 & 33.22/.9529 & 33.32/.9541 & 32.87/.9521 & \textcolor{red}{36.56}/\textcolor{red}{.9787} \\
         &  \multicolumn{1}{|c|}{4} & 30.79/.9343 & \textcolor{blue}{34.36}/\textcolor{blue}{.9612} & 33.16/.9530 & 32.74/.9499 & 33.19/.9545 & 32.99/.9526 & 33.40/.9553 & 32.61/.9497 & 33.13/.9532 & 33.21/.9543 & 32.70/.9521 & \textcolor{red}{36.40}/\textcolor{red}{.9778} \\

        \hline
        \hline
        \multirow{8}*{\rotatebox{90}{\textbf{HCIold} \cite{HCIold}}}
        &  \multicolumn{1}{|c|}{-4} & 36.85/.9775 & \textcolor{blue}{40.85}/\textcolor{blue}{.9875} & 39.36/.9852 & 38.44/.9833 & 39.18/.9852 & 39.22/.9851 & 39.55/.9858 & 38.69/.9837 & 38.93/.9849 & 39.20/.9851 & 39.17/.9850 & \textcolor{red}{42.34}/\textcolor{red}{.9929} \\
        &  \multicolumn{1}{|c|}{-3} & 36.83/.9775 & \textcolor{blue}{40.88}/\textcolor{blue}{.9874} & 39.57/.9854 & 38.45/.9837 & 39.35/.9854 & 39.33/.9853 & 39.76/.9858 & 38.99/.9843 & 39.18/.9850 & 39.37/.9851 & 39.40/.9852 & \textcolor{red}{43.04}/\textcolor{red}{.9936} \\
        &  \multicolumn{1}{|c|}{-2} & 36.84/.9777 & \textcolor{blue}{40.32}/.9871 & 38.84/.9858 & 38.05/.9841 & 38.33/.9854 & 38.80/.9852 & 38.70/.9862 & 38.64/.9851 & 38.90/.9860 & 38.47/.9855 & 39.53/\textcolor{blue}{.9879} & \textcolor{red}{42.80}/\textcolor{red}{.9938} \\
        &  \multicolumn{1}{|c|}{-1} & 36.71/.9782 & 40.22/.9873 & 40.43/.9902 & 39.44/.9891 & 39.60/.9900 & 39.79/.9895 & 40.96/.9914 & 39.68/.9899 & 41.19/.9915 & 40.73/.9913 & \textcolor{blue}{41.45}/\textcolor{blue}{.9923} & \textcolor{red}{43.31}/\textcolor{red}{.9952} \\
         &  \multicolumn{1}{|c|}{0} 
         & 37.69/.9785 & 41.05/.9875 & 43.42/.9932 & 43.81/.9938 
         & 44.20/.9942 & 44.45/.9946 & 44.23/.9941 
         & 44.08/.9942 
         & 44.84/.9948 
         & 44.52/.9945 
         & \textcolor{blue}{44.94}/\textcolor{red}{.9949} 
         & \textcolor{red}{45.08}/\textcolor{red}{.9949} \\
         &  \multicolumn{1}{|c|}{1} & 36.66/.9783 & 39.25/.9869 & 39.85/.9903 & 40.31/.9904 & 38.42/.9901 & 39.93/.9903 & 40.18/.9915 & 39.85/.9905 & \textcolor{blue}{40.88}/.9921 & 39.99/.9916 & 40.50/\textcolor{blue}{.9922} & \textcolor{red}{42.75}/\textcolor{red}{.9942} \\
         &  \multicolumn{1}{|c|}{2} & 36.74/.9779 & \textcolor{blue}{39.78}/\textcolor{blue}{.9871} & 38.77/.9862 & 38.50/.9844 & 38.25/.9862 & 38.70/.9856 & 38.41/.9865 & 38.17/.9847 & 38.64/.9861 & 38.61/.9867 & 38.33/.9863 & \textcolor{red}{42.31}/\textcolor{red}{.9939} \\
         &  \multicolumn{1}{|c|}{3} & 36.76/.9777 & \textcolor{blue}{40.66}/\textcolor{blue}{.9876} & 39.31/.9852 & 38.48/.9834 & 39.10/.9855 & 39.10/.9852 & 39.45/.9858 & 38.37/.9832 & 39.00/.9851 & 39.19/.9853 & 38.90/.9849 & \textcolor{red}{42.97}/\textcolor{red}{.9939} \\
         &  \multicolumn{1}{|c|}{4} & 36.80/.9776 & \textcolor{blue}{40.70}/\textcolor{blue}{.9877} & 39.21/.9853 & 38.68/.9838 & 39.03/.9855 & 39.09/.9853 & 39.35/.9859 & 38.36/.9834 & 38.68/.9848 & 39.00/.9851 & 38.68/.9848 & \textcolor{red}{42.67}/\textcolor{red}{.9935} \\
         
        \hline
        \hline
        \multirow{8}*{\rotatebox{90}{\textbf{INRIA} \cite{INRIA}}}
        &  \multicolumn{1}{|c|}{-4} & 31.58/.9566 & \textcolor{blue}{35.40}/\textcolor{blue}{.9769} & 34.24/.9719 & 33.75/.9695 & 34.42/.9725 & 33.99/.9713 & 34.64/.9736 & 33.89/.9703 & 34.13/.9719 & 34.37/.9724 & 34.20/.9720 & \textcolor{red}{36.46}/\textcolor{red}{.9815} \\
        &  \multicolumn{1}{|c|}{-3} & 31.55/.9566 & \textcolor{blue}{35.39}/\textcolor{blue}{.9768} & 34.22/.9717 & 33.71/.9695 & 34.43/.9726 & 34.04/.9715 & 34.62/.9736 & 33.95/.9703 & 34.12/.9715 & 34.39/.9726 & 34.10/.9710 & \textcolor{red}{36.67}/\textcolor{red}{.9826} \\
        &  \multicolumn{1}{|c|}{-2} & 31.55/.9567 & \textcolor{blue}{35.22}/\textcolor{blue}{.9763} & 34.04/.9715 & 33.59/.9695 & 34.08/.9718 & 33.87/.9709 & 34.31/.9726 & 33.91/.9707 & 34.13/.9721 & 34.11/.9716 & 34.67/.9749 & \textcolor{red}{36.67}/\textcolor{red}{.9829} \\
        &  \multicolumn{1}{|c|}{-1} & 31.49/.9573 & 35.26/.9767 & 34.88/.9767 & 34.59/.9760 & 34.92/.9770 & 34.56/.9757 & 35.51/.9790 & 34.69/.9766 & 35.42/.9790 & 35.26/.9783 & \textcolor{blue}{35.55}/\textcolor{blue}{.9799} & \textcolor{red}{36.79}/\textcolor{red}{.9837} \\
         &  \multicolumn{1}{|c|}{0} 
         & 31.33/.9577 & 35.01/.9769 & 35.39/.9804 & 35.28/.9832 
         & 36.15/.9842 & 35.80/.9843 & 36.36/.9840 
         & 36.09/.9849 
         & 36.57/.9853 
         & \textcolor{blue}{36.59}/\textcolor{blue}{.9855} 
         & \textcolor{blue}{36.59}/\textcolor{red}{.9859} 
         & \textcolor{red}{36.67}/.9853 \\
         &  \multicolumn{1}{|c|}{1} & 31.53/.9573 & 35.04/.9762 & 34.82/.9765 & 34.83/.9768 & 34.56/.9772 & 34.73/.9772 & 35.44/.9793 & 34.93/.9773 & \textcolor{blue}{35.30}/.9782 & 35.21/.9784 & 35.25/\textcolor{blue}{.9795} & \textcolor{red}{36.80}/\textcolor{red}{.9840} \\
         &  \multicolumn{1}{|c|}{2} & 31.55/.9567 & \textcolor{blue}{35.29}/\textcolor{blue}{.9765} & 34.16/.9721 & 33.75/.9698 & 34.43/.9735 & 33.99/.9717 & 34.49/.9737 & 33.75/.9706 & 34.07/.9720 & 34.46/.9740 & 34.08/.9726 & \textcolor{red}{36.75}/\textcolor{red}{.9832} \\
         &  \multicolumn{1}{|c|}{3} & 31.56/.9565 & \textcolor{blue}{35.41}/\textcolor{blue}{.9768} & 34.10/.9710 & 33.65/.9689 & 34.39/.9725 & 33.94/.9711 & 34.54/.9732 & 33.61/.9687 & 34.02/.9712 & 34.37/.9725 & 34.04/.9715 & \textcolor{red}{36.75}/\textcolor{red}{.9829} \\
         &  \multicolumn{1}{|c|}{4} & 31.58/.9565 & \textcolor{blue}{35.43}/\textcolor{blue}{.9769} & 34.18/.9715 & 33.80/.9696 & 34.40/.9724 & 34.01/.9713 & 34.63/.9736 & 33.72/.9695 & 34.03/.9713 & 34.36/.9723 & 34.02/.9715 & \textcolor{red}{36.59}/\textcolor{red}{.9821} \\

        \hline
        \hline
        \multirow{8}*{\rotatebox{90}{\textbf{STFgantry} \cite{STFgantry}}}
        &  \multicolumn{1}{|c|}{-4} & 29.83/.9479 & \textcolor{blue}{35.69}/\textcolor{blue}{.9833} & 33.73/.9739 & 32.48/.9677 & 34.19/.9776 & 32.92/.9715 & 34.70/.9792 & 32.98/.9702 & 33.87/.9751 & 34.11/.9775 & 33.58/.9751 & \textcolor{red}{39.33}/\textcolor{red}{.9947} \\
        &  \multicolumn{1}{|c|}{-3} & 29.80/.9479 & \textcolor{blue}{35.79}/\textcolor{blue}{.9832} & 33.78/.9740 & 32.59/.9688 & 34.44/.9781 & 33.12/.9723 & 34.78/.9794 & 33.25/.9714 & 33.92/.9750 & 34.34/.9778 & 33.89/.9755 & \textcolor{red}{39.68}/\textcolor{red}{.9950} \\
        &  \multicolumn{1}{|c|}{-2} & 29.82/.9484 & \textcolor{blue}{35.65}/\textcolor{blue}{.9831} & 33.83/.9769 & 32.59/.9716 & 33.70/.9789 & 32.56/.9734 & 34.26/.9808 & 33.39/.9754 & 34.31/.9793 & 33.84/.9792 & 34.05/.9821 & \textcolor{red}{39.43}/\textcolor{red}{.9950} \\
        &  \multicolumn{1}{|c|}{-1} & 29.72/.9490 & 35.44/.9830 & 35.56/.9860 & 34.37/.9837 & 35.89/.9881 & 34.09/.9831 & 36.46/.9890 & 34.89/.9860 & 36.53/.9895 & 36.34/.9895 & \textcolor{blue}{36.65}/\textcolor{blue}{.9903} & \textcolor{red}{39.65}/\textcolor{red}{.9952} \\
         &  \multicolumn{1}{|c|}{0} 
         & 31.06/.9498 & 36.33/.9831 & 38.36/.9904 & 37.95/.9898 
         & 39.64/.9929 & 38.72/.9909 & 39.61/.9926 
         & 38.77/.9915 
         & 39.86/.9936 
         & \textcolor{blue}{40.54}/.9941 
         & 40.40/\textcolor{blue}{.9942} 
         & \textcolor{red}{42.17}/\textcolor{red}{.9957} \\
         &  \multicolumn{1}{|c|}{1} & 29.72/.9490 & 34.87/.9830 & 34.97/.9862 & 34.67/.9846 & 34.64/.9890 & 34.10/.9851 & 35.60/.9902 & 34.96/.9862 & \textcolor{blue}{35.78}/.9893 & 35.66/\textcolor{blue}{.9906} & 35.15/.9901 & \textcolor{red}{38.81}/\textcolor{red}{.9949} \\
         &  \multicolumn{1}{|c|}{2} & 29.79/.9483 & \textcolor{blue}{35.01}/\textcolor{blue}{.9829} & 33.66/.9779 & 32.88/.9721 & 33.85/.9821 & 32.61/.9740 & 33.85/.9816 & 32.90/.9750 & 33.97/.9800 & 34.15/.9827 & 32.70/.9798 & \textcolor{red}{38.58}/\textcolor{red}{.9947} \\
         &  \multicolumn{1}{|c|}{3} & 29.77/.9477 & \textcolor{blue}{35.20}/\textcolor{blue}{.9831} & 33.45/.9731 & 32.50/.9676 & 33.96/.9779 & 32.90/.9715 & 34.18/.9787 & 32.41/.9683 & 33.53/.9743 & 33.94/.9777 & 33.02/.9741 & \textcolor{red}{38.53}/\textcolor{red}{.9949} \\
         &  \multicolumn{1}{|c|}{4} & 29.80/.9477 & \textcolor{blue}{35.19}/\textcolor{blue}{.9832} & 33.39/.9733 & 32.53/.9679 & 33.72/.9774 & 32.78/.9714 & 34.18/.9792 & 32.41/.9685 & 33.43/.9745 & 33.76/.9773 & 32.78/.9739 & \textcolor{red}{38.46}/\textcolor{red}{.9947} \\

        \hline
    \end{tabular}}
    \label{tab:SheardDisparity}
\end{table*}

\section{LF Angular SR}
\label{sec:LFASR}

It is worth noting that the proposed spatial-angular correlation learning mechanism has large potential in multiple LF image processing tasks. 
In this section, we apply our proposed spatial-angular correlation learning mechanism to the LF angular SR task. 
We first introduce our EPIT-ASR model for LF angular SR.
Then, we introduce the datasets and implementation details in our experiments. 
Finally, we present the preliminary but promising results as compared to the state-of-the-art LF angular SR methods.

\subsection{Upsampling}

Since our EPIT is flexible to LFs with different angular resolutions (as demonstrated in Sec.~\ref{sec:DiffAng}), the EPIT-ASR model can be built by changing the upsampling stage of EPIT. 

Here, we follow \cite{LF-Distg, FS_GAF} to take the 2$\times$2 $\rightarrow$ 7$\times$7 angular SR task as an example to introduce the angular upsampling module in our EPIT-ASR.
Given the deep LF feature ${\boldsymbol F} \in \mathbb{R}^{2 \times 2 \times H \times W \times C}$, a 2$\times$2 convolution without padding is first applied to the angular dimensions to generate an angular-downsampled feature ${\boldsymbol F}_{down} \in \mathbb{R}^{1 \times 1 \times H \times W \times C}$. 
Then, a 1$\times$1 convolution is used to increase the channel dimension, followed by a 2D pixel-shuffling layer to generate the angular-upsampled feature ${\boldsymbol F}_{up} \in \mathbb{R}^{7 \times 7 \times H \times W \times C}$. 
Finally, a 3$\times$3 convolution is applied to the spatial dimensions of ${\boldsymbol F}_{up}$ to generate the final output ${\mathcal L}_{RE} \in \mathbb{R}^{7 \times 7 \times H \times W}$.

\subsection{Datasets and Implement Details}
Following \cite{FS_GAF, LF-Distg}, we conducted experiments on the HCInew \cite{HCInew} and  HCIold \cite{HCIold} datasets. 
All LFs in these datasets have an angular resolution of 9$\times$9. 
We cropped the central 7$\times$7 SAIs with 64$\times$64 spatial resolution as groundtruth high angular resolution LFs, and selected the corner 2$\times$2 SAIs as inputs. 

Our EPIT-ASR was initialized using the Xavier algorithm \cite{Xavier}, and trained using the Adam method \cite{Adam} with $\beta_{1}=0.9$, $\beta_{2}=0.999$. 
The initial learning rate was set to 2$\times 10^{-4}$ and halved after every 15 epochs. 
The training was stopped after 80 epochs.
During the training phase, we performed random horizontal flipping, vertical flipping, and 90-degree rotation to augment the training data.

\subsection{Qualitative Results}

Figure~\ref{fig:reconstruction} shows the quantitative and qualitative results achieved by different LF angular SR methods. 
It can be observed that the magnitude of errors for our EPIT-ASR is smaller than other methods, especially on the delicate texture areas (e.g., the letters in scene {\texttt{Dishes}}). 
As shown in the zoom-in regions, our method generates more faithful details with fewer artifacts.

\begin{figure*}[t]
    \centering
    \includegraphics[width=17.2cm]{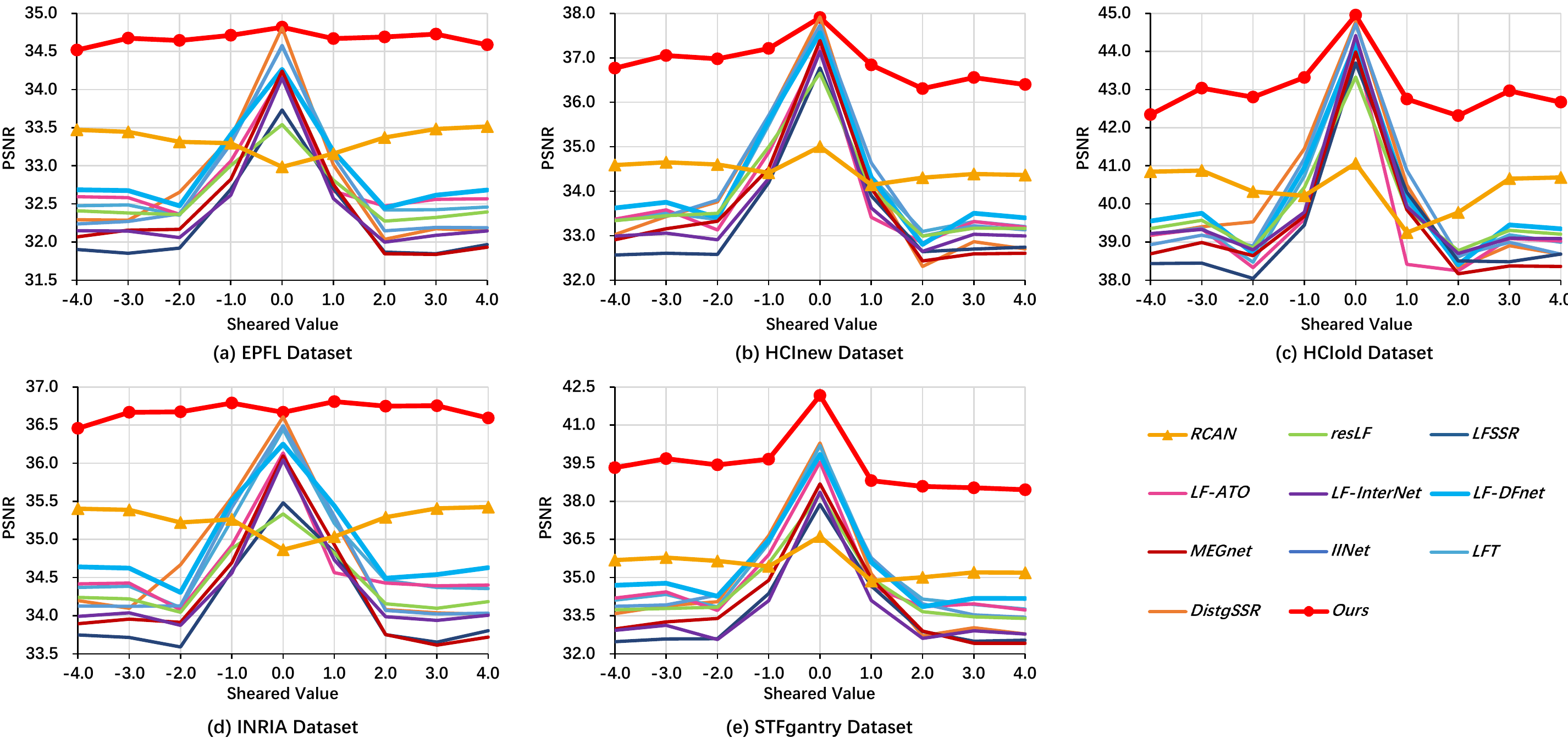}
    \caption{Quantitative comparison of different SR methods on five datasets with different shearing values for 2$\times$ SR. }
    \label{fig:shearedDisparity_x2}
\end{figure*}

\begin{figure*}[t]
    \centering
    \includegraphics[width=17.2cm]{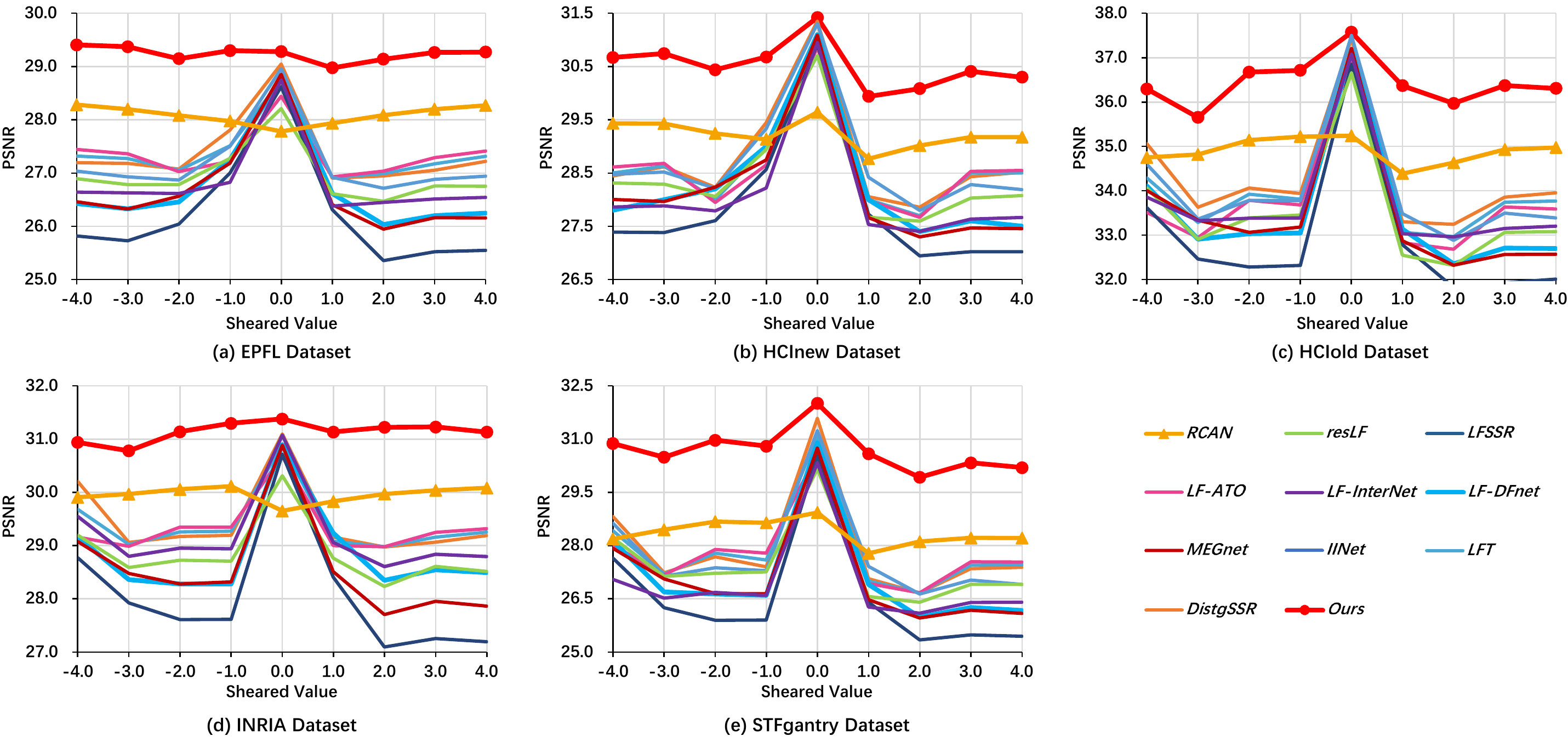}
    \caption{Quantitative comparison of different SR methods on five datasets with different shearing values for 4$\times$ SR. }
    \label{fig:shearedDisparity_x4}
\end{figure*}

\begin{figure*}[t]
    \centering
    \includegraphics[width=17.4cm]{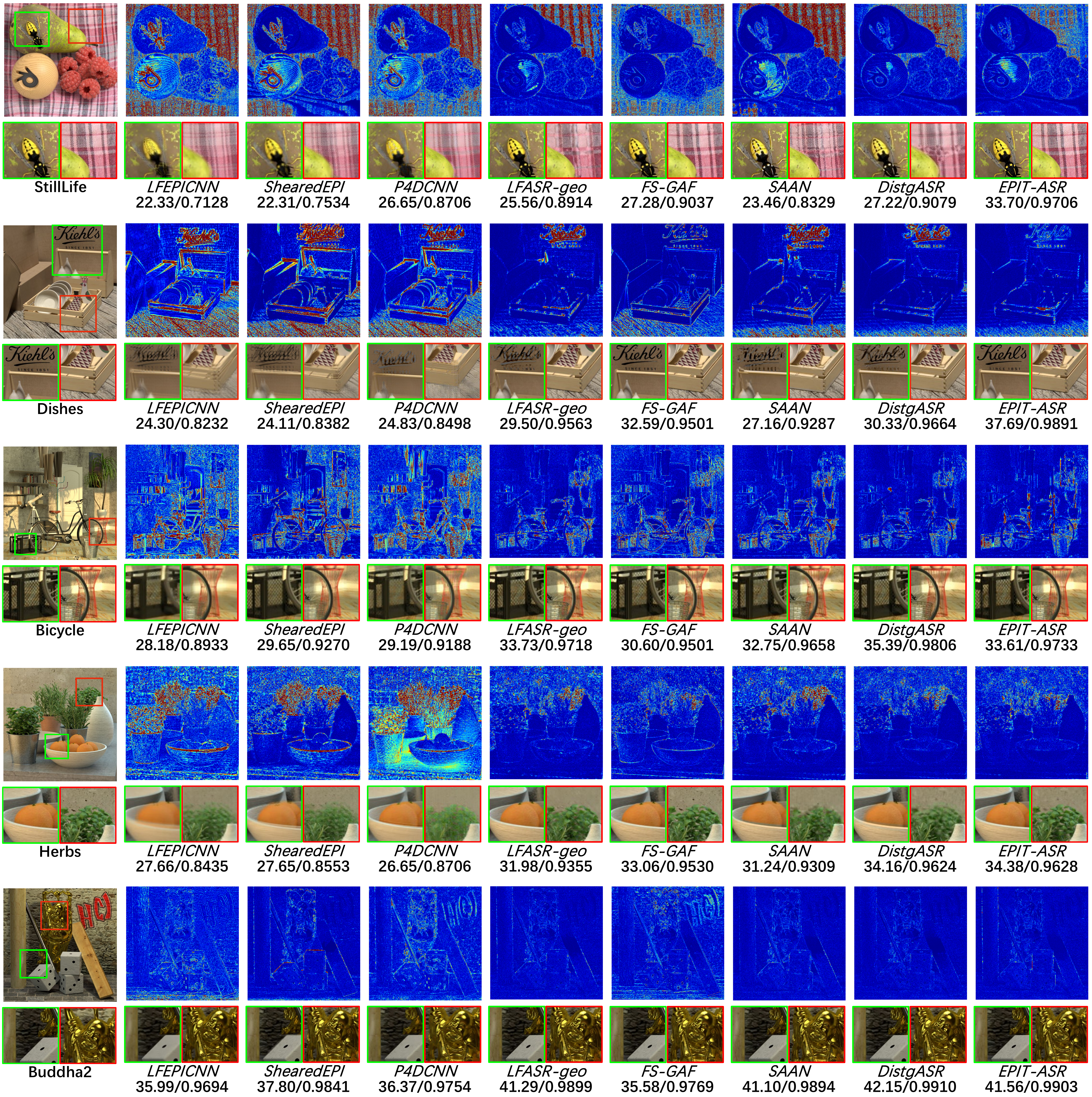}
    \caption{
    Visual results achieved by different methods on scenes {\texttt{StillLife}}, {\texttt{Dishes}}, {\texttt{Bicycle}}, {\texttt{Herbs}} and {\texttt{Buddha2}} for 2$\times$2 $\rightarrow$ 7$\times$7 angular SR. Here, we show the error maps of the reconstructed center view images, along with two zoom-in regions for qualitative comparison. The PSNR and SSIM values achieved on each scene are reported for quantitative comparison. Zoom in for the best view.
    }
    \label{fig:reconstruction}
\end{figure*}

}

\end{document}